\documentclass[10pt,twocolumn,letterpaper]{article}

\usepackage[pagenumbers]{wacv} 

\usepackage{graphicx}
\usepackage{amsmath}
\usepackage{amssymb}
\usepackage{booktabs}

\usepackage[numbers,sort,compress]{natbib}
\usepackage{soul}
\usepackage[dvipsnames]{xcolor}
\usepackage{xspace}
\usepackage[normalem]{ulem}
\usepackage{shadowtext}

\usepackage{tabularx,ragged2e,booktabs}
\usepackage{multirow}

\usepackage{color, colortbl}
\definecolor{Gray}{gray}{0.9}

\usepackage{enumitem}

\usepackage[heightadjust=all,valign=c]{floatrow}
\newfloatcommand{capbtabbox}{table}[][\FBwidth]

\usepackage[outline]{contour}
\usepackage{mathtools}
\usepackage{bbm}

\usepackage{sg-macros} %

\makeatletter
\renewcommand{\paragraph}{%
  \@startsection{paragraph}{4}%
  {\z@}{1.05ex \@plus 1ex \@minus .2ex}{-1em}%
  {\normalfont\normalsize\bfseries}%
}
\makeatother

\usepackage[pagebackref,breaklinks,colorlinks]{hyperref}

\DeclareRobustCommand{\textblueb}[1]{{\begingroup{\color{blue}\textbf{#1}}\endgroup}}

\begin{document}

\title{Bi-directional Training for Composed Image Retrieval via\\Text Prompt Learning}

\author{Zheyuan Liu$^{1}$ \enskip Weixuan Sun$^{1}$ \enskip Yicong Hong$^{1}$ \enskip Damien Teney$^{2,3}$ \enskip Stephen Gould$^{1}$ \\
$^{1}$Australian National University\\
$^{2}$Idiap Research Institute\enskip
$^{3}$Australian Institute for Machine Learning, University of Adelaide \\
{\tt\small \{zheyuan.liu, weixuan.sun, stephen.gould\}@anu.edu.au} \\
{\tt\small mr.yiconghong@gmail.com, damien.teney@idiap.ch}
}

\maketitle

\begin{abstract}
	Composed image retrieval searches for a target image based on a multi-modal user query comprised of a reference image and modification text describing the desired changes.
	Existing approaches to solving this challenging task learn a mapping from the (reference image, modification text)-pair to an image embedding that is then matched against a large image corpus.
	One area that has not yet been explored is the reverse direction, which asks the question, what reference image when modified as described by the text would produce the given target image?
	In this work we propose a bi-directional training scheme that leverages such reversed queries and can be applied to existing composed image retrieval architectures with minimum changes, which improves the performance of the model.
	To encode the bi-directional query we prepend a learnable token to the modification text that designates the direction of the query and then finetune the parameters of the text embedding module.
	We make no other changes to the network architecture.
	Experiments on two standard datasets show that our novel approach achieves improved performance over a baseline BLIP-based model that itself already achieves competitive performance.
  Our code is released at \href{https://github.com/Cuberick-Orion/Bi-Blip4CIR}{\textit{https://github.com/Cuberick-Orion/Bi-Blip4CIR}}.
\end{abstract}


\section{Introduction}

Composed image retrieval (CIR)~\cite{Vo_2019_tirg,fashioniq,Liu:CIRR} aims at retrieving images based on the user input of a reference image and a sentence stating certain desired changes. 
The retrieved target image must encompass the user-specified changes while remaining similar to the reference image on other aspects.
Unlike conventional image~\cite{10.1145/500141.500159_cbir} or text-based~\cite{zhang_tbir,6126478_tbir} retrieval with one input modality, CIR is more precise, which is appreciated in domains such as image search or e-commercial product retrieval.

\begin{figure}[t]
  \begin{center}
    \includegraphics[trim={0 5pt 0 0},clip, width=0.95\linewidth]{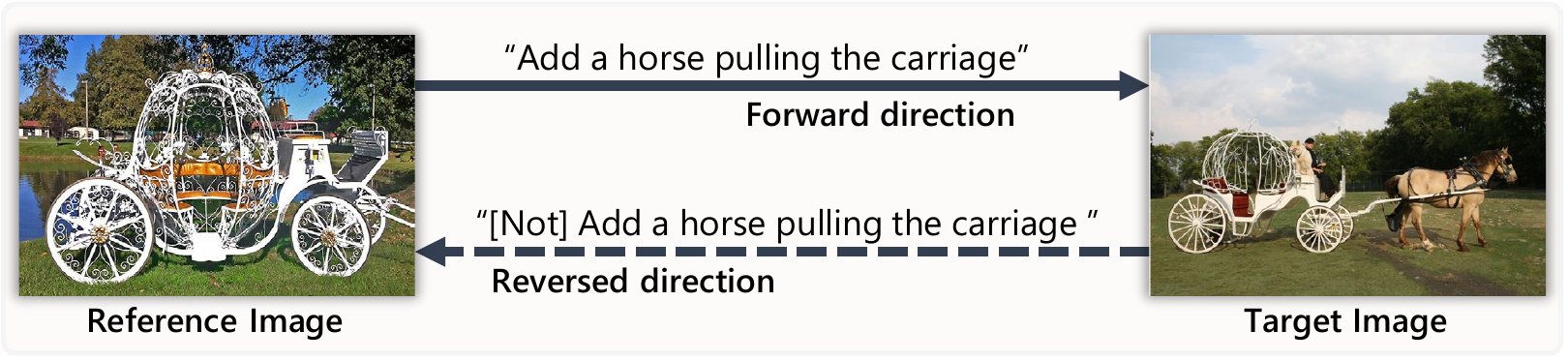}
  \end{center}
  \caption{\textbf{Forward direction:} Existing methods on composed image retrieval focus on mapping the input $\langle$reference image, modification text$\rangle$-pair to the target image. \textbf{Reversed direction:} We propose to also exploit information in the mapping from the $\langle$target image, reversed modification text$\rangle$-pair to the reference image.
  Note that the reversed text is an illustration, which we do not have and have to infer from existing data.}
  \label{fig:intro-0}
  \vspace{-12pt}
\end{figure}

Existing approaches~\cite{Vo_2019_tirg,chen2020image_val,dodds2020modality_maaf,Anwaar2020CompositionalLO} on CIR mostly focus on learning the mapping of embeddings from the given $\langle$reference image, text$\rangle$-pair to a target image, where the multi-modal input pair is jointly embedded through various mechanisms including gating~\cite{Vo_2019_tirg}, multi-layer attention~\cite{chen2020image_val,dodds2020modality_maaf} or convex summation~\cite{Baldrati_2022_CVPR_clip4cir0}. 
However, few have considered the full potential of the information available in the training data. 
If we view conventional CIR methods as training in the forward direction from reference to target, while conditioning on the text; then it is easy to see that a reverse direction can be achieved from target to reference, provided that a suitable text reversal can be found (\figref{fig:intro-0}). 
Harnessing information in such a bi-directional fashion benefits the training and increases the robustness of the model.
We argue that this is particularly valuable given the limited dataset sizes~\cite{fashioniq,Liu:CIRR} and high annotation cost~\cite{Pic2Word} of existing datasets for this task.

\begin{figure*}[ht!]
  \begin{center}
    \includegraphics[trim={0pt 0pt 0 0pt},clip, width=0.99\linewidth]{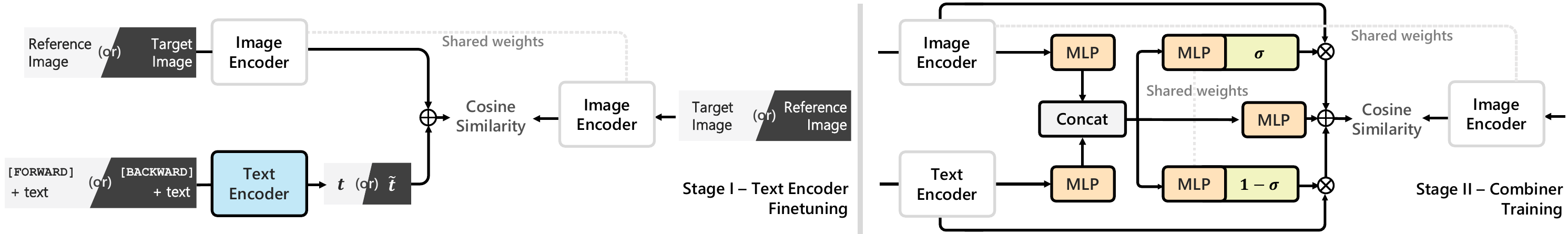}
  \end{center}
  \caption{Our bi-directional training pipeline. 
  Only modules with colors are being updated.
  Here, $\otimes$ denotes element-wise product and $\oplus$ denotes element-wise addition.
  Note the details of contrastive loss is not fully shown here (see~\secref{sec:bi-directional-training}), instead, we only illustrate with the positive target/reference images.
  \textbf{Left:} Text encoder finetuning, where we input queries of both directions (distinguished by light/dark colors), and infer $\tilde{t}$ through training, along with $t$. Note the bi-directional tokens prepended to the text.
  We omit the \texttt{[CLS]} token from the text encoder for brevity.
  \textbf{Right:} Combiner training detailing the architecture~\cite{Baldrati_2022_CVPR_clip4cir1}.
  We train the model with bi-directional queries simultaneously, as in the previous stage.
  }\label{fig:model-0}
  \vspace{-12pt}
\end{figure*}

To this end, we propose bi-directional training for CIR to simultaneously learn to retrieve images (target or reference, respectively) from both the forward and reversed queries.
As stated above, one major challenge is text reversal, that is, to create text that carries the opposite semantic meanings compared to the original.
For human annotators, rewriting the text can be easy. 
However, in the absence of additional annotations, the naive approach of reversing text through handcrafted linguistic rules can be difficult. 
This is especially true for the CIRR dataset on generic images, as the sentences are of great variations and high complexity.
One approach is to pass the text embeddings through a dedicated module (\eg an MLP) when inversion is needed, while preserving the rest of the joint embedding model and train end-to-end.
We argue a better alternative is to leverage the text encoders of vision-and-language pretrained (VLP) models, which are commonly used in CIR to extract text embeddings. The advantage of which is that such encoders are pretrained on massive corpus, and, hence, are powerful in capturing the nuanced semantics of natural language.
Specifically, we take inspiration from recent work on few-shot text-guided image editing, \ie text inversion~\cite{ruiz2022dreambooth} and propose to prepend the text with special learnable tokens signifying its directionality.
We then leverage the state-of-the-art two-stage training pipeline~\cite{Baldrati_2022_CVPR_clip4cir1} of the combiner architecture~\cite{Baldrati_2022_CVPR_clip4cir0}, where the first stage is to finetune the text encoder.
We discover that through finetuning, the text encoder can associate the concept of directionality of the text with the special text tokens, and generate different embeddings from the same text input.
This allows us to effectively treat queries of both directions equally after finetuning, and make no further changes to the subsequent joint embedding model (\ie the combiner) or its training process when we include the reversed queries.
In order to include the reversed queries in training, we involve a secondary loss term in the contrastive loss. To better take advantage of the reverse queries we employ a modified sampling strategy for negative samples, such that the loss is more coherent in the bi-directional training scheme, as will be detailed in~\secref{sec:bi-directional-training}.

In summary, we propose a bi-directional training scheme for composed image retrieval (CIR), which jointly trains on the forward queries, \ie from reference to target images, as well as the reversed queries from target to reference.
To obtain text embeddings of reversed semantics, we prepend the text with learnable tokens and finetune the text encoders. 
Additionally, we modify the contrastive loss on the reversed path. 
No further changes to the model architecture or training pipeline are needed, which makes our approach easily applicable to existing methods.
We empirically show that our approach achieves improved performance on datasets of diverse domains over a BLIP-based baseline that has already achieved state-of-the-art performance.

\section{Related Work}

\paragraph{Composed Image Retrieval.}
The task of composed image retrieval (CIR) introduced by \citet{Vo_2019_tirg} aims at studying the composition of multi-modal features, where initially inputs of low complexity~\cite{johnson2017clevr,StatesAndTransformations_MITstates} are considered. It is later adopted for fashion product~\cite{fashioniq,han2017automatic_fashion200k,10.5555/1886063.1886114_shoes}, and recently further extend into generic images~\cite{Liu:CIRR}.
Most existing methods follow a fusion paradigm, where the features of the input reference image and text are jointly embedded and compared against features of all candidate target images for the closest match. 
Extensive research~\cite{Vo_2019_tirg,chen2020image_val,dodds2020modality_maaf,Anwaar2020CompositionalLO,Baldrati_2022_CVPR_clip4cir0,Liu:CIRR} is done on the fusion mechanism of the network, with the recent state-of-the-art~\cite{Baldrati_2022_CVPR_clip4cir0,Baldrati_2022_CVPR_clip4cir1} adopting a combiner architecture that performs a convex combination of the input modalities within the CLIP~\cite{CLIP} feature manifold.
We note that this is the first method that simultaneously achieves state-of-the-art on both fashion and generic image datasets.
Meanwhile, in the contemporary unpublished work, \citet{levy2023dataroaming} and \citet{liu2023candidate} both adopt the BLIP~\cite{li2022blip} multimodal encoder that further improves the performance. 
Concurrently, others have explored splitting CIR into two-stage with coarse and fine searching~\cite{Yu2020CurlingNetCL}, disentangling the image-image and image-text matching into dual branches~\cite{ARTEMIS}, and expanding the task into zero-shot scenarios~\cite{Pic2Word}.
Among methods developed for CIR, we are mostly related to DCNet~\cite{kim:2021:AAAI_dcnet}, where a correction module is used to model the difference between the reference and target images and match it to the text. In essence, it explores a different directionality of the training data than ours.
Compared to DCNet, our method does not require an additional module, or joint loss that connects the said module to the main network. 
Instead, our bi-directional training treats samples of both directions in the same manner.
We also avoid computing feature differences through subtractions, which can be hard to learn~\cite{LASO}.

\paragraph{Cycle consistency.}
Conceptually, our bi-directional training resembles the cycle-consistency concept seen in vision-and-language (\eg Robust VQA~\cite{meet2019cycle_vqa}) and generation (\eg CycleGAN~\cite{CycleGAN2017}) tasks, as we share the philosophy of manipulating model inputs and outputs to further exploit information in the training instances. 
However, such a concept bears different motivations and designs under various task setups. 
For VQA, it is implemented as a secondary question-answering stage with generated rephrases of the question, which improves the robustness of the model under linguistic variations.
CycleGAN, however, utilizes cycle consistency with the absence of paired training instances between two domains.
We note that CIR is different in that, three entities are included in the input and output. This setup requires unique designs, hence, making our method fundamentally different from previous work. 

\paragraph{Text inversion.}
Recent work~\cite{ruiz2022dreambooth,gal2022textual} on few-shot text-guided image editing shows that it is possible to bind the appearance of a certain instance in an image with artificially injected special tokens within the text prompt through finetuning, so that the model can generate diverse samples containing said instance. A technique termed text inversion.
Though not entirely equivalent, we take inspiration from the above work and adopt a similar idea in finetuning the text encoder. Our intention is to associate the concept of text directionality with special tokens, so that the model can recognize the need of reversing the semantics of the language.

\section{Bi-directional Composed Image Retrieval}

Given the embeddings of a query of $\langle$reference image, modification text$\rangle$-pair denoted as $q=\langle I_\text{R}, t\rangle$, the objective is to locate a target image that best matches the query, whose embedding is denoted as  $I_\text{T}$.
Our goal is to also learn on the reversed query $\tilde{q}=\langle I_\text{T}, \tilde{t}\rangle$ simultaneously, which maps from the target $I_\text{T}$ to the $I_\text{R}$. Here, $\tilde{t}$ represents the text embedding that is semantically reversed. However, $\tilde{t}$ is not directly computable as the text associated with such reversed embeddings do not exist in the training data.
To overcome this difficulty we propose to infer the semantically reversed text embedding $\tilde{t}$ from the original modification text using the text encoder.

\begin{figure}[t!]
  \begin{center}
    \includegraphics[trim={0pt 0pt 0 0pt},clip, width=0.95\linewidth]{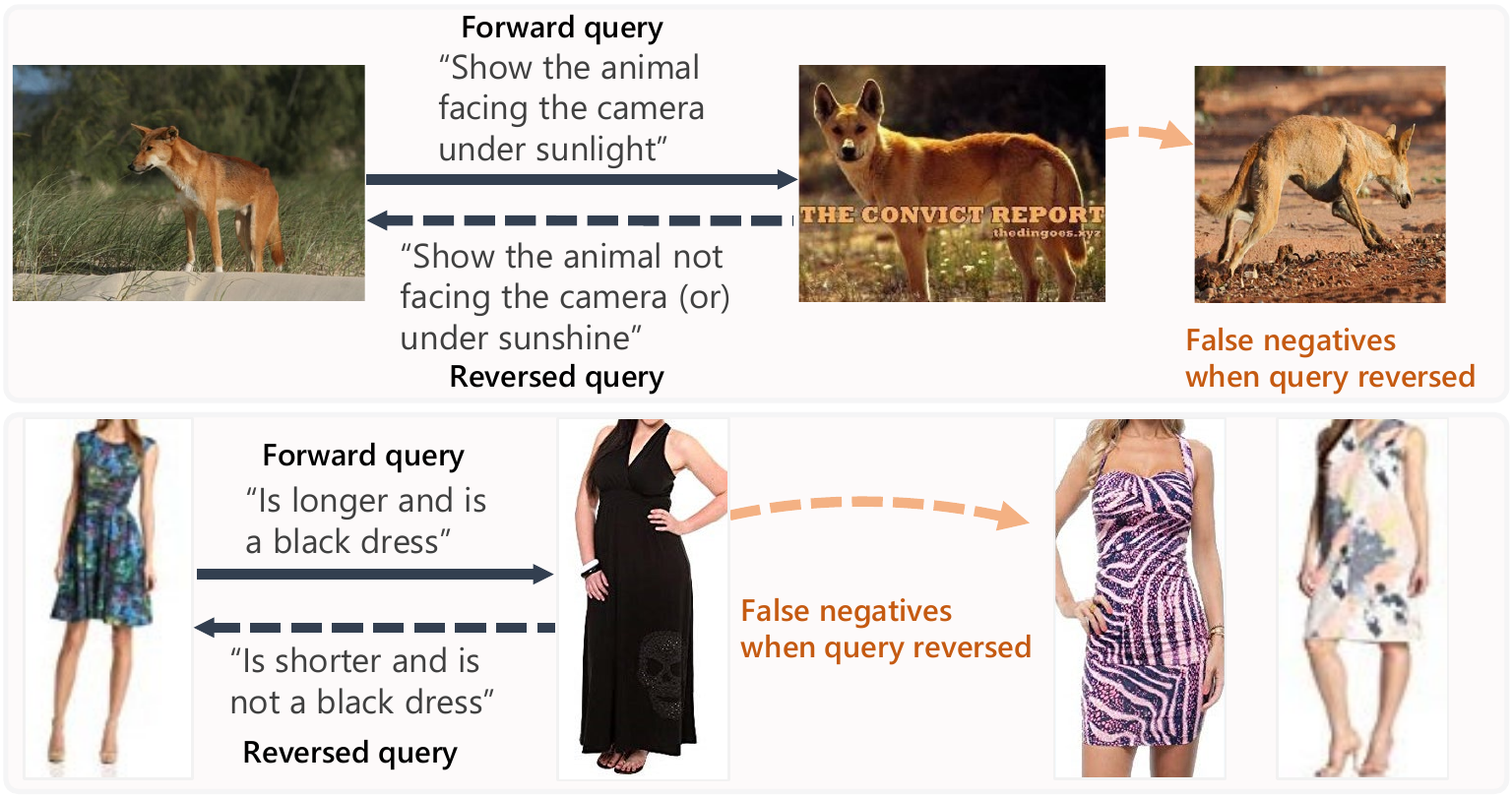}
  \end{center}
  \caption{False negatives can exist in the reversed direction. 
  A one-to-one text in the forward direction can become one-to-many when its semantics are reversed.
  Since we lack human labels for the reversed queries, only the original reference image is deemed positive.
  We point out that this issue is related to both the semantics of the text, as well as the image corpus. Hence, the prevalence of false negatives might vary for individual samples in different datasets.
  Note that the reversed text here is provided to illustrate the case, we do not have such annotations in the datasets.
  }\label{fig:false-neg}
  \vspace{-12pt}
\end{figure}

Our method is based on a recent architecture by~\citet{Baldrati_2022_CVPR_clip4cir1}, which is a two-stage design as illustrated in~\figref{fig:model-0}. Though, as an augmentation scheme other models are also applicable, provided that they utilize tokenized text embeddings that is common in VLP models. 
In the following subsections, we first propose our main idea on leveraging the text prompts for inferring the reverse query (\secref{sec:bi-directional-training}). We then detail the model and training pipeline adapted from \cite{Baldrati_2022_CVPR_clip4cir0,Baldrati_2022_CVPR_clip4cir1} (\secref{sec:model-arch-train-pipeline}). Finally, we discuss the false negatives in the reversed queries in \secref{sec:false-neg}, which impacts our inference strategy.

\subsection{Bi-directional Training}\label{sec:bi-directional-training}

\paragraph{Text Prompt Learning.}
As shown in~\figref{fig:model-0} (left), we leverage the first-stage text encoder finetuning to infer $\tilde{t}$ alongside $t$, such that it can produce text embedding of either direction for a given text.
Specifically, we prepend special learnable tokens to the modification text sentences. The idea is to bind the concept of query directionality to such tokens through learning.

Here one option is to make no changes to the text of the forward queries, and only inject a token when the text needs to be reversed. The implication is that the forward queries shall be trained in their de-facto forms (see \secref{sec:model-arch-train-pipeline}), while only making necessary changes to the additional augmentations included.
However, this introduces an asymmetry in the treatment of the forward and reversed queries.
Our intuition is that the balanced approach will force the model to better recognize the purpose of the injected tokens and distinguish between the forward and reversed modes.

We choose \texttt{[FORWARD]} and \texttt{[BACKWARD]} as the learnable tokens for the forward and reversed queries respectively in all our experiments. 
Together with the \texttt{[CLS]} token from the text encoder, which is pretrained to summarize the text, a tokenized text sequence $\mathfrak{t}$ of $\{\mathfrak{t_1} \cdots \mathfrak{t_n}\}$ in the forward query is then processed into $\{ \texttt{[CLS]}, \texttt{[FORWARD]}, \mathfrak{t_1} \cdots \mathfrak{t_n} \}$, which is passed to the text encoder for embedding. Likewise for the reversed case with $\{\mathfrak{t_1} \cdots \mathfrak{t_n} \}$ unmodified.

We note that prompt design~\cite{lester-etal-2021-power,ruiz2022dreambooth} could potentially be of significant value to the end results. 
However, our focus is on the bi-directional training scheme, hence, we choose to rely on simple, generic tokens and make no specific adjustments to suit each dataset.

\paragraph{Modifications to the Training Pipeline.}
A favorable characteristic of our approach is the minimum changes made to the existing training pipeline.
Here, we leverage the state-of-the-art two-stage training scheme (details of which are in~\secref{sec:model-arch-train-pipeline}), as illustrated in~\figref{fig:model-0}.
We note that, after finetuning the text encoder to infer $\tilde{t}$, the reversed queries $\langle q=\langle I_\text{T}, \tilde{t}\rangle, I_\text{R}\rangle$ can be constructed by a simple exchange of image orders.
The entire process is computed on-the-fly with low additional cost.
We then treat all queries equally regardless of their directionality, which allows us to train the second-stage combiner with bi-directional samples simultaneously without changes to its architecture.

\paragraph{Negative Sampling in Contrastive Loss.}
We follow previous work and use the batch-based classification (BBC) loss~\cite{Vo_2019_tirg}.
Given a batch size of $B$, with the embeddings of the $i$-th query pair $\langle I_\text{R}^i, t^i\rangle$, its corresponding positive target $I_\text{T}^i$, the forward query loss is computed as:
\begin{equation}\label{eq:loss-f}
\resizebox{.87\hsize}{!}{$
  \mathcal{L}_\text{F} = -\frac{1}{B} \displaystyle \sum_{i=1}^B \log\!\left[\vcenter{\hbox{$\displaystyle \frac{ \exp \Bigl[ \lambda \cdot \kappa \left( f(I_\text{R}^i, t^i), I_\text{T}^i \right) \Bigr] }{ \displaystyle \sum_{j=1}^B \exp \Bigl[ \lambda \cdot \kappa \left( f(I_\text{R}^i, t^i), I_\text{T}^j \right) \Bigr] } $}}\right]$
},
\end{equation}
where $f(\cdot,\cdot)$ denotes the combination function, $\kappa(\cdot,\cdot)$ is the similarity kernel implemented as cosine similarity and $\lambda$ is the temperature parameter. We follow~\citet{Baldrati_2022_CVPR_clip4cir1} and set $\lambda$ to 100 in all our experiments.
As the denominator shows, we normalize over all other matches within a batch in training, which includes both the positive $I_\text{T}^i$ and all other target images in the batch as negatives $I_\text{T}^j$ for $j \neq i$.

When training on the reversed queries that maps query $q=\langle I_\text{T}, \tilde{t}\rangle$ to reference image $I_\text{R}$, multiple options exist in sampling the negatives to form a contrastive loss.
Here, we propose to formulate the loss as:
\begin{equation}\label{eq:loss-b-modified}
  \resizebox{.87\hsize}{!}{$
  \mathcal{L}_\text{B} = -\frac{1}{B} \displaystyle \sum_{i=1}^B \log\!\left[\vcenter{\hbox{$\displaystyle \frac{ \exp \Bigl[ \lambda \cdot \kappa \left( f(I_\text{T}^i, \tilde{t}^i), I_\text{R}^i \right) \Bigr] }{ \displaystyle \sum_{j=1}^B \exp \Bigl[ \lambda \cdot \kappa \left( f(I_\text{T}^j, \tilde{t}^i), I_\text{R}^i \right) \Bigr] } $}}\right]$
},
\end{equation}
where we have chosen to sample the negatives among candidate target images, \ie $I_\text{T}^j$, as in \eqnref{eq:loss-f}.
Our intuition is to unify losses for the forward and reversed queries so that they are learned to contrast against the same group of negatives. 
We empirically confirm that such a loss obtains better performance compared to, \eg sampling negatives among the $I_\text{R}^j$ (see \secref{sec:ablation-loss}).

The loss for bi-directional training is then computed as the weighted sum of forward and reversed terms,
\begin{equation}\label{eq:loss-total}
  \mathcal{L} = \mathcal{L}_\text{F} + \alpha \mathcal{L}_\text{B},
\end{equation}
where $\alpha$ is a hyperparameter to balance the magnitudes of the two loss terms.
We refer readers to~\secref{sec:balancing-losses} for details on determining this parameter and analysis.

\subsection{Model Architecture and Training Pipeline}\label{sec:model-arch-train-pipeline}
We base our bi-directional training on a recent state-of-the-art baseline obtained using the combiner architecture~\cite{Baldrati_2022_CVPR_clip4cir0,Baldrati_2022_CVPR_clip4cir1} with BLIP~\cite{li2022blip} vision-and-language pretrained (VLP) model, termed BLIP4CIR.
We follow~\citet{Baldrati_2022_CVPR_clip4cir1} and adopt a two-stage training scheme as follows.

\paragraph{Text Encoder Finetuning.}
As shown in \figref{fig:model-0} (left), we first finetune the text encoder.
The architecture is relatively light with the multi-modal combination done through an element-wise addition. The output is compared against embeddings of candidate targets through cosine similarity.
Note that the image encoder is kept frozen as it is prohibitively expensive to finetune.

The intuition of this finetuning stage is to reduce the domain gap between the pretraining tasks and the downstream task, CIR. 
Conceptually, the element-wise addition is used to encourage the output $t$ as a displacement vector between $I_\text{R}$ and $I_\text{T}$ in the image manifold.
Once the text encoder is finetuned, we freeze it and train a combiner module that replaces the element-wise addition mentioned above, which involves more sophisticated joint embedding functions.

\paragraph{Combiner Training.}
In the second-stage training, combiner replaces element-wise addition as the joint embedding function, with architecture depicted in \figref{fig:model-0} (right).
We refer readers to~\cite{Baldrati_2022_CVPR_clip4cir0} for details.
Overall, it can be viewed as a three-branch summation, with two branches implemented as a convex combination of the two input modalities, and a third branch of a learned image-text mixture.
During combiner training, both the image and text encoders are frozen.

\paragraph{BLIP Embeddings.}
We propose to change the CLIP~\cite{CLIP} image and text encoders in both stages to BLIP~\cite{li2022blip}.
The motivation is two-fold. 
First, BLIP has been demonstrated to be a powerful VLP network, and through early experiments, we found its text encoder to be better than CLIP's.
Second, we notice that BLIP is trained on text of potentially higher complexity that better matches the annotations in CIR datasets. 
In contrast, while the training data of CLIP is proprietary, the qualitative examples demonstrated by~\citet{CLIP} are less sophisticated in general. 
We hypothesize that the BLIP text encoder is of stronger reasoning ability, and is, therefore, better suited to CIR.
This is especially true for bi-directional training, as the text encoder needs to reverse the semantics of a given sentence.
We empirically confirm that BLIP provides a much stronger baseline over the stock CLIP encoders.
The change from pre-trained CLIP to BLIP encoders is straightforward, as they are of similar transformer~\cite{vaswani2017attention_transformer} structures. 
Note that we do not involve the cross-attention fusion layer in BLIP that jointly embeds image and text modalities. 
Instead, we treat the image and text encoders as separate modules, as shown in \figref{fig:model-0}. This is to align with the usage of CLIP so that we can preserve the stock training pipeline.

\subsection{False Negatives in Reversed Queries}\label{sec:false-neg}

A systematic issue for our bi-direction training scheme is the occurrences of false negatives in the reversed queries.
As illustrated in \figref{fig:false-neg}, a one-to-one modification text in the forward query could become one-to-many when its semantics are reversed. 
For instance, a text of ``\emph{change to a white cat}'' will be reversed to ``\emph{change \textbf{from} a white cat}'', which now leads to all cats (gray, ginger, fawn, etc.) that are not white --- yet, only the original reference image is labeled as the positive.
We especially note that the prevalence of false negatives in the reversed path is related to the semantics of the text as well as the image corpus (\eg following the example above, the number cat images in the corpus shall affect the number of false negatives). Hence, the scale of such an issue can vary for individual samples across datasets.
This renders global label smoothing techniques~\cite{labelsmoothing,Sun_2023_CVPR} ineffective in our testings, as it could inadvertently affect queries with few false negatives.

Granted, false negatives also exist in the forward path~\cite{fashioniq,Liu:CIRR}, due to the prohibitive high cost in exhaustively labeling all candidates for each query.
But they are less common, as human annotations are often sufficiently specific. Plus, existing metrics are designed to mitigate the issue by using Recall@$K$ of large $K$ values.

We argue that nevertheless, training on the reversed queries still benefits the model, as demonstrated by our experiments of improved performance. However, special attention shall be paid in balancing the magnitudes of the loss terms (\eqnref{eq:loss-total}), as the loss of the reversed path is generally higher.
We also note that the prevalence of false negatives in the reversed direction suggests that validating on such reversed queries will lead to inferior results (see~\secref{sec:inference-on-reversed-queries} for details).

\paragraph{Inference Strategy.}
Following the above, we design our inference strategy mirroring existing work~\cite{Vo_2019_tirg,Baldrati_2022_CVPR_clip4cir1}, which only performs on the forward queries.
For each query, we rank the similarities of the combined $(I_\text{R},t)$ embedding with all candidate $I_\text{T}'$ and pick the highest as the prediction.

\section{Experiments and Discussions}

\paragraph{Datasets and Metrics.}
We follow \citet{Baldrati_2022_CVPR_clip4cir1} and test on two datasets of different domains.

\textbf{Fashion-IQ}~\cite{fashioniq} focuses on fashion products of three subtypes, \textit{Dress, Shirt} and \textit{Toptee}. In total it contains over 30k triplets sampled from 77k images.
Each triplet includes two human-generated annotations.
We follow previous work and report Recall@$K$ with $K=10$ and $50$, while comparing the overall model performance with $(\text{R@}10+\text{R@}50)/2$, as advised by~\citet{fashioniq}. The choice of $K$ accounts for the potential false negatives in the forward queries. All results are on the validation set, as the test set is not publicly available.

\textbf{CIRR}~\cite{Liu:CIRR} includes around 36k triplets sampled from 21k generic images sourced from NLVR$^2$~\cite{Suhr_2019_nlvr2}. The human annotations are of higher complexity compared to Fashion-IQ.
The dataset is designed to overcome the issue of false negatives, as it is prohibitive to exhaustively label every candidate target for each input query. 
Specifically,~\citet{Liu:CIRR} group images by subsets of six and draw reference-target pairs from them. When annotating a given pair, annotators are instructed to avoid creating false negatives within the subset from which the pair is drawn. 
To this end, the evaluation protocol for CIRR is designed to be a combination of standard Recall@$K$ with $K=$1, 5, 10, 50 and Recall$_\text{Subset}$@$K$ with $K=$1, 2, 3, where Recall$_\text{Subset}$@$K$ only considers candidates from the same subset as the pair.
Following~\citet{Liu:CIRR}, we assess the overall model performance with $(\text{R@}5+\text{R}_\text{Subset}\text{@}1)/2$. Results in the main table are reported on the test set, the ground truths of which are not available. Instead, we obtain results through the official evaluation server\footnote{\href{https://cirr.cecs.anu.edu.au/}{\textit{https://cirr.cecs.anu.edu.au/}}}. Results of ablation studies are reported on the validation set.

\paragraph{Implementation Details.}\label{sec:implement-detail}
We adopt the default image pre-processing scheme and model configuration as in~\cite{Baldrati_2022_CVPR_clip4cir1}, except when BLIP encoders~\cite{li2022blip} require a different setting in dimensions. This includes the input image resolutions ($384\times 384$) and the combiner input feature dimension ($256$ from BLIP encoder outputs).
For finetuning the text encoder, we follow BLIP downstream task settings and optimize with AdamW~\cite{Loshchilov2017DecoupledWD_adamw} for 15 epochs, with a learning rate of $5 \times 10^{-5}$, a weight decay of 0.05 and a cosine learning rate schedule.
We increase the learning rate of the last linear projection layer to $5 \times 10^{-3}$ to speed up the convergence.
For training the combiner, we adopt the original settings detailed in~\cite{Baldrati_2022_CVPR_clip4cir0} and train for 200 epochs.
The batch size of baseline experiments (\ie without bi-directional training) follows~\cite{Baldrati_2022_CVPR_clip4cir1}. 
The batch size of all bi-directional training experiments is reduced by half due to the GPU memory limit.

All experiments are trained with mixed-precision~\cite{https://doi.org/10.48550/arxiv.1710.03740_mixedprecision} in PyTorch with one NVIDIA A100 80G. We base our implementation on the official codebases released by~\citet{Baldrati_2022_CVPR_clip4cir1}~\footnote{\href{https://github.com/ABaldrati/CLIP4Cir}{\textit{https://github.com/ABaldrati/CLIP4Cir}}} and~\citet{li2022blip}~\footnote{\href{https://github.com/salesforce/BLIP}{\textit{https://github.com/salesforce/BLIP}}}.

\begin{table*}[th!]
  \centering \scalebox{0.66}{
  \begin{tabular}{p{0.03\linewidth}p{0.30\linewidth}rrrrrrrrr} 
  \toprule
  \multicolumn{1}{c}{} & \multicolumn{1}{c}{} & \multicolumn{2}{c}{\textbf{Dress}} & \multicolumn{2}{c}{\textbf{Shirt}} &\multicolumn{2}{c}{\textbf{Toptee}} &\multicolumn{2}{c}{\textbf{Average}} & \textbf{Avg.} \\
  \cmidrule(lr){3-4}
  \cmidrule(lr){5-6}
  \cmidrule(lr){7-8}
  \cmidrule(lr){9-10}
  \multicolumn{1}{l}{} & \multicolumn{1}{l}{\textbf{Methods}} & R@10 & R@50 & R@10 & R@50 & R@10 & R@50 & R@10 & R@50 & \textbf{Metric}  \\ 
  \midrule
  \textbf{1} & MRN~\cite{MRN}    &12.32    & 32.18          & 15.88 & 34.33           & 18.11  & 36.33    & 15.44 & 34.28 & 24.86    \\
  \textbf{2} & FiLM~\cite{perez2017film}    &14.23    & 33.34          & 15.04 & 34.09           & 17.30  & 37.68    & 15.52 & 35.04 & 25.28    \\
  \textbf{3} & TIRG~\cite{Vo_2019_tirg}    &14.87    & 34.66          & 18.26 & 37.89           & 19.08  & 39.62    & 17.40 & 37.39 & 27.40    \\
  \textbf{4} & Relationship~\cite{santoro2017simple}    &15.44    & 38.08          & 18.33 & 38.63     & 21.10  & 44.77    & 18.29 & 40.49 & 29.39    \\
  \textbf{5} & CIRPLANT~\cite{Liu:CIRR}    &14.38   & 34.66     & 13.64 & 33.56     & 16.44  & 38.34   & 14.82 & 35.52 & 25.17  \\ 
  \textbf{6} & CIRPLANT~w/OSCAR~\cite{Liu:CIRR} &17.45   & 40.41     & 17.53 & 38.81     & 21.64  & 45.38   & 18.87 & 41.53 & 30.20   \\ 
  \textbf{7} & VAL~w/GloVe~\cite{chen2020image_val}       &22.53    & 44.00          & 22.38 & 44.15     & 27.53  & 51.68    & 24.15 & 46.61 & 35.40    \\
  \textbf{8} & CurlingNet~\cite{Yu2020CurlingNetCL} & 24.44  &  47.69    & 18.59 & 40.57  &  25.19 &  49.66  & 22.74 &  45.97 &  34.36  \\ 
  \textbf{9} & DCNet~\cite{kim:2021:AAAI_dcnet} & 28.95 & 56.07  &  23.95 & 47.30  & 30.44 & 58.29 & 27.78 & 53.89 &  40.84  \\ 
  \textbf{10} & CoSMo~\cite{Lee2021CoSMoCM} & 25.64 & 50.30 & 24.90 & 49.18 & 29.21 & 57.46 &26.58&52.31&  39.45  \\ 
  \textbf{11} & MAAF~\cite{dodds2020modality_maaf} &23.8\phantom{0}    & 48.6\phantom{0}          & 21.3\phantom{0} & 44.2\phantom{0}     & 27.9\phantom{0}  & 53.6\phantom{0}    & 24.3\phantom{0} & 48.8\phantom{0} & 36.6\phantom{0}    \\
  \textbf{12} & ARTEMIS~\cite{ARTEMIS} &  25.68 & 51.25   & 28.59 & 55.06 &   21.57 & 44.13    & 25.25 & 50.08   &  37.67  \\ 
  \textbf{13} & SAC~w/BERT~\cite{SAC} &26.52 & 51.01 & 28.02 & 51.86 & 32.70 & 61.23 &29.08  &  54.70 & 41.89 \\ 
  \textbf{14} & AMC~\cite{AMC} & 31.73  &  59.25    & 30.67 &  59.08    & 36.21  & 66.06   & 32.87 & 61.64 & 47.25   \\ 
  \textbf{15} & CLIP4CIR~\cite{Baldrati_2022_CVPR_clip4cir1} &33.81   & 59.40     & 39.99 & 60.45     & 41.41  & 65.37   & 38.32 & 61.74 & 50.03   \\ 
  \specialrule{\lightrulewidth}{0pt}{1pt}
  \textbf{16} & CASE$^\dagger$~\cite{levy2023dataroaming} & \textbf{47.77}  & \textbf{69.36}     & \textbf{48.48} &  \textbf{70.23}   & \textbf{50.18} & \textbf{72.24}  &  \textbf{48.79} & \textbf{70.68} & \textbf{59.74}  \\
  \specialrule{\lightrulewidth}{0pt}{1pt}
  \rowcolor{Gray}
  \textbf{17} & BLIP4CIR~(first-stage) &37.13   & 62.67     & 35.92 & 60.40     & 43.60  & 68.28   & 38.88 & 63.78 & 51.33   \\ 
  \textbf{18} & BLIP4CIR & 40.65  & 66.34    & 40.38 &  64.13    & \textblueb{46.86} & 69.91  & 42.63 & 66.79 & 54.71   \\ 
  \rowcolor{Gray}
  \textbf{19} & BLIP4CIR+Bi~(first-stage) &36.94   & 63.71     & 37.49 & 60.06     & 43.60  & 67.77   & 39.34 & 63.85 & 51.60   \\ 
  \textbf{20} & BLIP4CIR+Bi & \textblueb{42.09}  & \textblueb{67.33}     & \textblueb{41.76} &  \textblueb{64.28}    & 46.61 & \textblueb{70.32}  & \textblueb{43.49}  & \textblueb{67.31} &  \textblueb{55.40} \\ 
  
  \bottomrule
  \end{tabular}}
  \caption{Comparison on Fashion-IQ validation set, we follow~\citet{fashioniq} to report Avg. Metric as \textit{(R@10+R@50)/2}. 
  $\dagger$~Contemporary work to ours, included for completeness.
  Best (resp. second-best) numbers are in bold-black (resp. blue), this excludes intermediate first-stage text encoder fine-tuning results marked in gray (rows 17, 19). 
  BLIP4CIR denotes the baseline using BLIP encoders. +Bi denotes the bi-directional training.
  For CLIP4CIR~\cite{Baldrati_2022_CVPR_clip4cir1}, we report their best-performing model that uses the two-stage training with RN50x4 as backbone. 
  Rows 1-2 are cited from~\cite{fashioniq}.
  Methods leveraging additional data and auxiliary tasks (\eg~\cite{han2023famevil}) are not included.
  }
  \label{tab:baseline_1}
\end{table*}
\begin{table*}[th!]
  \centering \scalebox{0.66}{
  \begin{tabular}{p{0.03\linewidth}p{0.34\linewidth}rrrrrrrr} 
  \toprule
  \multicolumn{1}{c}{} & \multicolumn{1}{c}{}        & \multicolumn{4}{c}{\textbf{Recall@}$\boldsymbol{K}$ }               & \multicolumn{3}{c}{\textbf{Recall}$_{\text{\textbf{Subset}}}$\textbf{@}$\boldsymbol{K}$ }   & \textbf{Avg.}       \\
  \cmidrule(lr){3-6}
  \cmidrule(lr){7-9}
  \multicolumn{1}{l}{} & \multicolumn{1}{l}{\textbf{Methods}} & $K=1$           & $K=5$           & $K=10$ & $K=50$          & $K=1$           & $K=2$           & $K=3$            &  \textbf{Metric} \\ 
  \midrule
  \textbf{1}  & TIRG~\cite{Vo_2019_tirg}              & 14.61  & 48.37  & 64.08  & 90.03 &  22.67  & 44.97  & 65.14  & 35.52 \\ 
  \textbf{2}  & TIRG$+$LastConv~\cite{Vo_2019_tirg}        & 11.04  & 35.68  & 51.27  & 83.29 &  23.82  & 45.65  & 64.55  & 29.75 \\ 
  \textbf{3} & MAAF~\cite{dodds2020modality_maaf}         & 10.31     & 33.03     & 48.30    & 80.06 &  21.05    & 41.81    & 61.60  & 27.04 \\
  \textbf{4} & MAAF$+$BERT~\cite{dodds2020modality_maaf}  & 10.12     & 33.10     & 48.01 & 80.57    & 22.04    & 42.41    & 62.14   & 27.57 \\
  \textbf{5} & MAAF$-$IT~\cite{dodds2020modality_maaf}    & 9.90      & 32.86     & 48.83 & 80.27    & 21.17    & 42.04    & 60.91   & 27.02 \\
  \textbf{6} & MAAF$-$RP~\cite{dodds2020modality_maaf}    & 10.22     & 33.32     & 48.68 & 81.84    & 21.41    & 42.17    & 61.60   & 27.37 \\
  \textbf{7} & CIRPLANT~\cite{Liu:CIRR} & 15.18  & 43.36  & 60.48  & 87.64 &  33.81  & 56.99  & 75.40  & 38.59 \\ 
  \textbf{8} & CIRPLANT~w/OSCAR~\cite{Liu:CIRR} & 19.55  & 52.55  & 68.39  & 92.38 &  39.20  & 63.03  & 79.49  & 45.88 \\ 
  \textbf{9} & ARTEMIS~\cite{ARTEMIS}    & 16.96     & 46.10     & 61.31 & 87.73    & 39.99    & 62.20    & 75.67   & 43.05 \\
  \textbf{10} & CLIP4CIR~\cite{Baldrati_2022_CVPR_clip4cir1}    & 38.53     & 69.98     & 81.86 & 95.93    & 68.19    & 85.64    & 94.17   & 69.09 \\
  \specialrule{\lightrulewidth}{0pt}{1pt}
  \textbf{11} & CASE$^\dagger$~\citep{levy2023dataroaming} & \textbf{48.00} & \textbf{79.11}  & \textbf{87.25}  & \textbf{97.57}  & \textbf{75.88}  & \textbf{90.58}   & \textbf{96.00}  & \textbf{77.50} \\
  \specialrule{\lightrulewidth}{0pt}{1pt}
  \rowcolor{Gray}
  \textbf{12} & BLIP4CIR~(first-stage)    & 35.18     & 67.11    & 79.18 & 94.70    & 68.71    & 86.65    & 94.51   & 67.90 \\
  \textbf{13} & BLIP4CIR  & \textblueb{40.17}  & 71.81   &  83.18   & 95.69  & \textblueb{72.34}    & \textblueb{88.70}   &  95.23  &  72.07  \\
  \rowcolor{Gray}
  \textbf{14} & BLIP4CIR+Bi~(first-stage)   & 35.30    & 67.42   & 79.88   &  94.58 & 68.55   & 86.46    & 94.75 & 67.99 \\
  \textbf{15} & BLIP4CIR+Bi & 40.15  & \textblueb{73.08}   & \textblueb{83.88}   & \textblueb{96.27} & 72.10   &  88.27   & \textblueb{95.93}  & \textblueb{72.59} \\
  \bottomrule
  \end{tabular}}
  \caption{Comparison on CIRR test set. We follow~\citet{Liu:CIRR} and report the Avg. Metric as \textit{(Recall@5+Recall}$_\text{Subset}$\textit{@1)/2}. 
  $\dagger$ Contemporary work to ours, included for completeness. Note that CASE can be additionally pre-trained on the large-scale LaSCo dataset~\cite{levy2023dataroaming}. Here, we include the results without such pre-training for a fair comparison.
  Best (resp. second-best) numbers are in bold-black (resp. blue), this excludes intermediate first-stage text encoder fine-tuning results marked in gray (rows 12, 14). 
  BLIP4CIR denotes the baseline using BLIP encoders. +Bi denotes the bi-directional training.
  For CLIP4CIR~\cite{Baldrati_2022_CVPR_clip4cir1}, we report their best-performing model that uses the two-stage training with RN50x4 as backbone. 
  Rows 1-8 are cited from~\cite{Liu:CIRR}.
  }\label{tab:baseline_0}
\end{table*}

\begin{table*}[tp]
  \centering \scalebox{0.66}{
  \begin{tabular}{p{0.03\linewidth}p{0.15\linewidth}ccrrrrrrr} 
  \toprule
   & & &         & \multicolumn{3}{c}{\textbf{Fashion-IQ}}               & \multicolumn{4}{c}{\textbf{CIRR}} \\
  \cmidrule(lr){5-7}
  \cmidrule(lr){8-11}
  \multicolumn{1}{l}{} &\textbf{Methods} & Neg-Sampling & Bi-Token  & R@10  & R@50  & Average & R@1 &  R@5   & R$_\text{Subset}$@1  & Average  \\ 
  \midrule
  \textbf{1}  & BLIP4CIR   & --- & --- & 42.63 & 66.79 & 54.71 & 41.11 & 74.89 & 72.66 & 73.78 \\ 
  \midrule
  \textbf{2}  & BLIP4CIR+Bi   & $\circ$  & & 42.43 & 67.23 & 54.83 & 40.61 & 74.46 & \textbf{73.36} & 73.91 \\ 
  \textbf{3}  & BLIP4CIR+Bi   & & $\circ$ & 43.30 & 66.77 & 55.03 & 40.28 & 73.95 & 72.57 & 73.26 \\ 
  \midrule
  \textbf{4}  & BLIP4CIR+Bi   &  \multicolumn{2}{c}{\textsc{Fully-Configured}}  & \textbf{43.49} & \textbf{67.31} & \textbf{55.40} & \textbf{42.36} & \textbf{75.46}  & 72.90 & \textbf{74.18} \\ 
  \bottomrule
  \end{tabular}}
  \caption{Ablation studies on Fashion-IQ and CIRR. Best numbers are in bold.
  Results reported on validation sets. 
  For Fashion-IQ, we report the average Recall@10 and 50 of all three categories. 
  For CIRR, the Average column denotes \textit{(Recall@5+Recall}$_\text{Subset}$\textit{@1)/2}, as in~\tabref{tab:baseline_0}.
  \textit{Bi-Token} suggests adding learnable tokens to queries of both directions (\ie bi-directional).
  \textit{Neg-Sampling} represents our negative sampling scheme in the reversed contrastive loss.
  Here $\circ$ denotes the item we are ablating (\ie where we remove this particular item from the fully-configured model and assess the outcome).
  }\label{tab:ablate_0}
\end{table*}

\subsection{Results on Fashion-IQ}\label{sec:results-fashion}

\tabref{tab:baseline_1} compares our approach with existing state-of-the-art methods on Fashion-IQ. We note that our BLIP-based baseline model (row 18) outperforms all previous approaches by a large margin, albeit exceeded by the contemporary work CASE (row 16), also using BLIP.
Impressively, through the first-stage text encoder finetuning (row 17), the performance already surpasses the previous CLIP-based best-performing model (row 15), demonstrating the high quality of BLIP embeddings.

With the addition of bi-directional training, the overall performance is further improved consistently throughout the two stages (rows 17 \textit{vs.} 19; rows 18 \textit{vs.} 20). On the final results obtained in the second stage in row 20, we gain notable improvements on categories of \textit{Dress} and \textit{Shirt}, while retaining approximately a similar performance on \textit{Toptee} compared to the BLIP baseline in row 18.
We conjecture that each sub-class might benefit from the bi-directional training differently, due to the quality of the reversed queries considering the issue of the false negatives and the specific image corpus (see~\secref{sec:false-neg}).

Our method is outperformed by the contemporary work CASE~\cite{levy2023dataroaming}.
We conjecture the main reason to be that the pre-trained BLIP~\cite{li2022blip} multimodal encoder adopted by CASE is more powerful than the combiner, which is also observed in its concurrent work~\cite{liu2023candidate} with further performance increase.
In comparison, we only leverage the BLIP visual and text encoders for feature extraction.
Notably, \citet{levy2023dataroaming} have introduced a similar bi-directional data augmentation scheme as ours on CASE that shows a benefit, effectively demonstrating the generalizability of this idea.

\subsection{Results on CIRR}\label{sec:exp-cirr}

\begin{figure}[tp]
  \begin{center}
    \begin{subfigure}{.49\textwidth}
      \centering
      \includegraphics[trim={10pt 0pt 43pt 32pt},clip, width=0.99\linewidth]{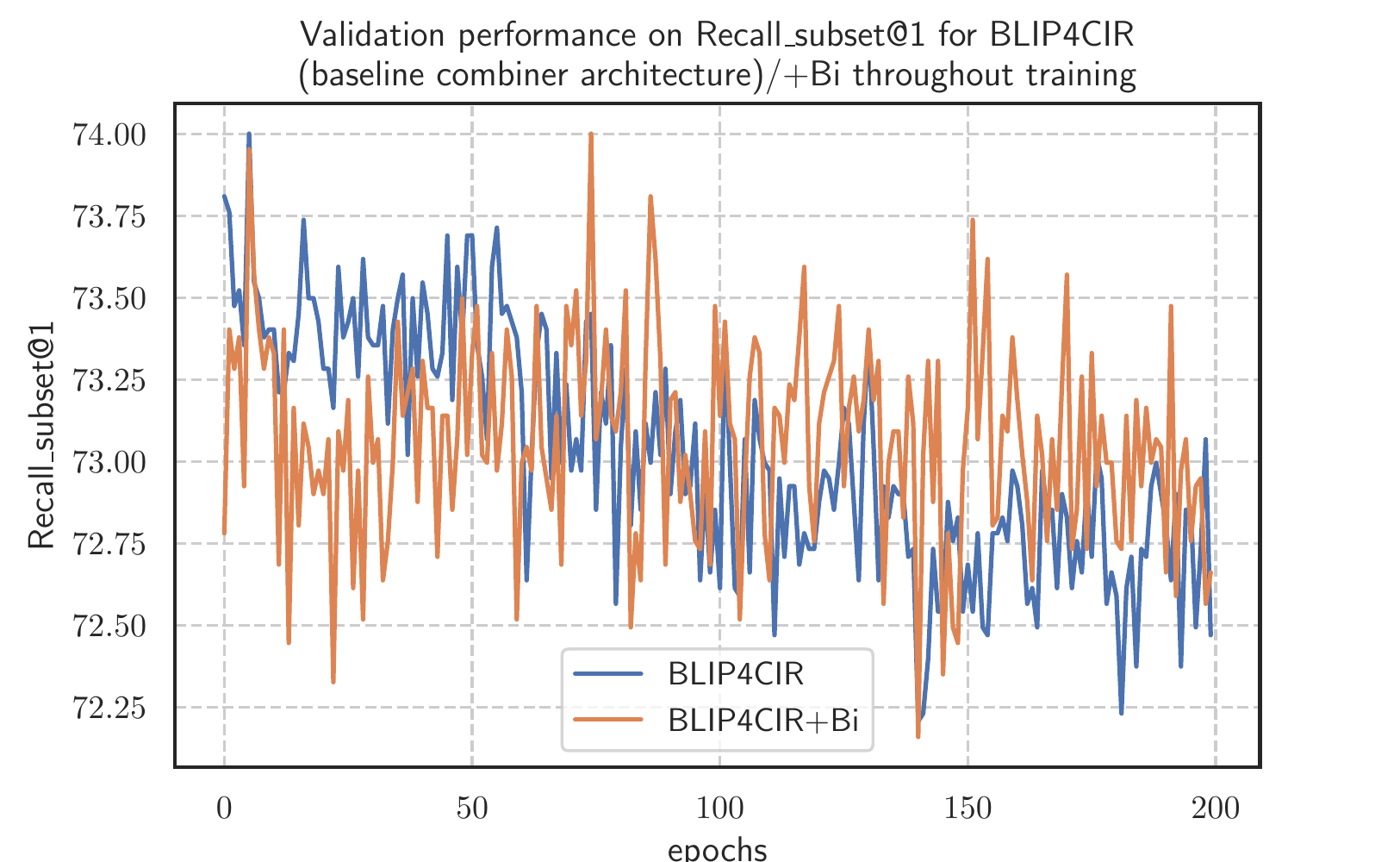}
    \end{subfigure}%
    \begin{subfigure}{.49\textwidth}
      \centering
      \includegraphics[trim={20pt 0pt 33pt 32pt},clip, width=0.99\linewidth]{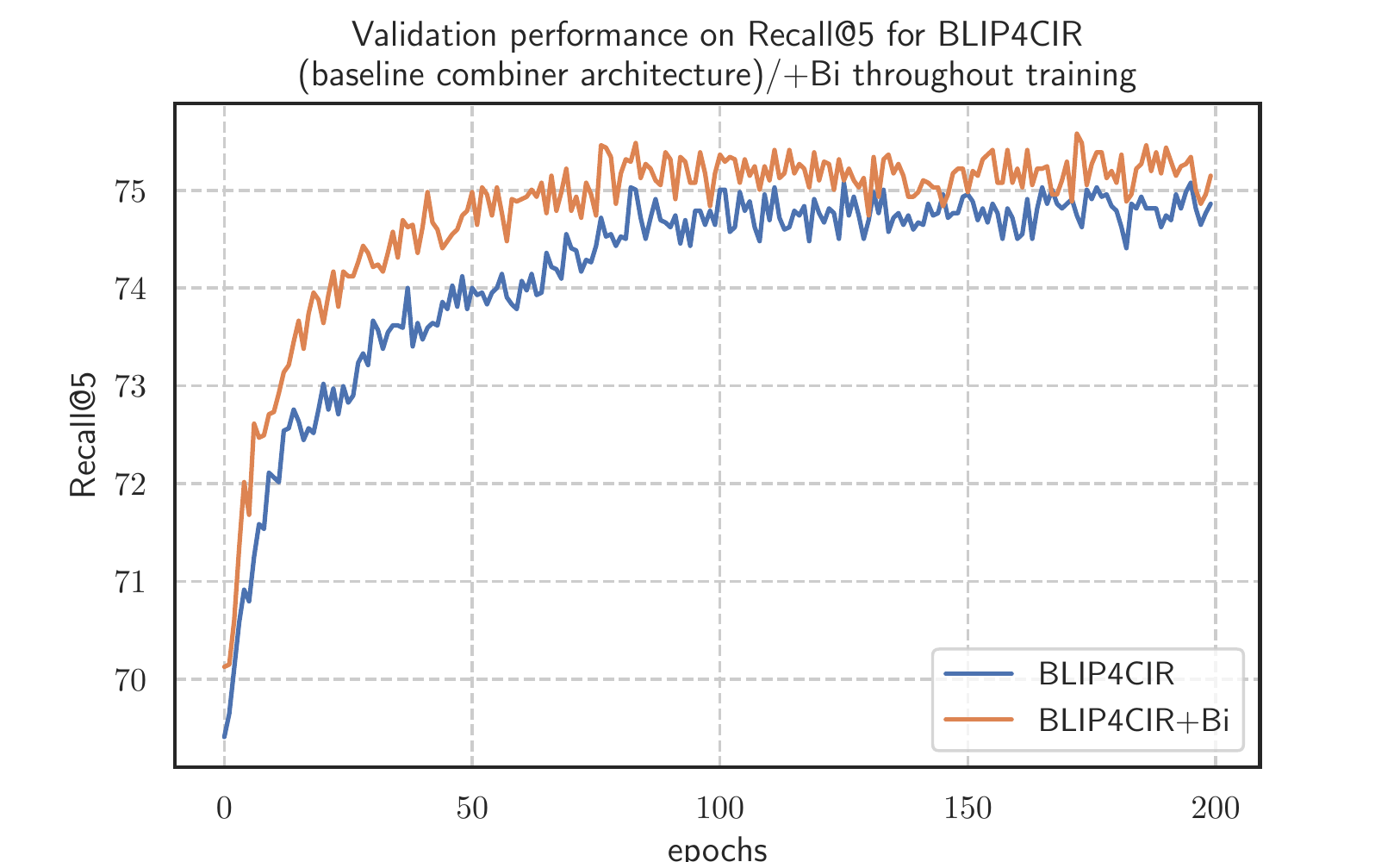}
    \end{subfigure}
  \end{center}
  \caption{
    Validation performance on \textbf{(left)} Recall$_\text{Subset}$@1 and \textbf{(right)} Recall@5 trained for 200 epochs. Results obtained on the CIRR validation set. \textbf{BLIP4CIR (blue):} baseline model with combiner architecture. \textbf{BLIP4CIR+Bi (orange):} combiner with our bi-directional training scheme.
  We observe that the combiner architecture gains very little, if none on the more challenging Recall$_\text{Subset}$ metric with high fluctuations among epochs. And that our method, as an augmentation scheme, inherits such a property.
  }\label{fig:val-recall-subset-1-vs-recall-1}
  \vspace{-16pt}
\end{figure}

\tabref{tab:baseline_0} compares the performance of state-of-the-art methods with our approach on CIRR test set. We note that BLIP embeddings (row 13) brings a consistent performance increase compared to the previous CLIP-based model (row 10), as in Fashion-IQ.
Our bi-directional training scheme (row 15) brings further increase to the performance of the strong baseline.
Admittedly, the performance increase is inconsistent across metrics, particularly for Recall$_\text{Subset}$. We conjecture two reasons that could account for this. 
First, the high complexity of the text in CIRR would render the learning of the reversed semantics harder.
Second, as demonstrated in~\figref{fig:val-recall-subset-1-vs-recall-1} (left, blue), we notice that the combiner gains very little, if none, on Recall$_\text{Subset}$ throughout training, and that the fluctuations in performance are high among epochs.
Given that Recall$_\text{Subset}$ is both more challenging by design and of high granularity, as it only considers five candidates from the same image subset, we point out the possibility that the state-of-the-art combiner architecture, though powerful, may fail in this metric.
To compare, we overlay the validation curve obtained using our bi-directional training scheme, as in~\figref{fig:val-recall-subset-1-vs-recall-1} (left, orange). We note that globally, our performance on Recall$_\text{Subset}$@1 is on par with the baseline. However, since we assess the performance via both the Recall@5 and Recall$_\text{Subset}$@1, the reported performance on the test set (\tabref{tab:baseline_0} row 15) is not necessarily optimal on each individual metric.
We stress that our bi-directional training, as a data augmentation scheme, does \textit{not} aim at improving the existing model architecture. It is therefore unsurprising that our result exhibits a similar learning behavior on Recall$_\text{Subset}$ as the combiner baseline.

In contrast, on Recall, our method gains noticeable improvements when $K=5$ and onwards in~\tabref{tab:baseline_0} (row 15 \textit{vs.} 13). We demonstrate that the performance improvement is consistently observed during training, as in~\figref{fig:val-recall-subset-1-vs-recall-1} (right, orange \textit{vs.} blue), suggesting valuable information exists in the reversed queries that benefits the learning. Regarding the similar performance on Recall@1 when compared with the baseline, we point to the fact that Recall@1 could potentially be impacted by false negatives in the forward queries~\citep{Liu:CIRR}. Given that CIRR is designed to assess Recall@$K$ with $K=5$, we conclude that our bi-directional training is generally beneficial to this task.

\subsection{Ablation Studies}
We conduct ablation studies on the two design choices of our method detailed in~\secref{sec:bi-directional-training}, as shown in \tabref{tab:ablate_0}.

\paragraph{Negative Sampling in Reversed Contrastive Loss.}\label{sec:ablation-loss}
In \tabref{tab:ablate_0} row 2, we compare the negative sampling scheme in the reversed path. Here, we report results obtained by contrasting against $I_\text{R}^j$ as opposed to $I_\text{T}^j$ in \eqnref{eq:loss-b-modified}.
We note that the proposed sampling scheme is vital to the bi-directional training on both datasets. Specifically, we find that if adopting the negative sampling scheme on $I_\text{R}^j$, the model gains very little from the bi-directional training, with Recall@$K$ either dropping below the baseline (row 1) or merely slightly improving over it, let alone performing on par with the fully-configured model (row 4). 
We conjecture that a negative sampling scheme on $I_\text{R}^j$ in the reverse queries causes a misalignment between the forward and reversed training paths, as the model is trained to contrast against $I_\text{T}^j$ in the forward queries. To compare, our proposed sampling technique is more coherent and yields better performance.
We note that the Recall$_\text{Subset}$@1 for CIRR presents as an outlier in this case. In this particular case, we propose to assess the performance difference more on the global Recall metrics, and refer to our discussions on the granularity of Recall$_\text{Subset}$ as well as the general learning behaviour of the baseline method in~\secref{sec:exp-cirr}.

\paragraph{Bi-directional Text Tokens.}\label{sec:ablation-bi-token}
\tabref{tab:ablate_0} row 3 demonstrates the effectiveness of adding learnable tokens in \textit{both} directions of the text rather than in the reversed direction alone. Interestingly, we discover that compared to Fashion-IQ, CIRR is more benefited from such a technique. Without it, the performance, in particular on Recall@5, drops significantly (row 3 \textit{vs.} 4).
The reason is thought to be the complexity of the text inputs. In Fashion-IQ, the text is often short and carries simple meanings (\figref{fig:qualitative-0} e-h), such as ``longer sleeves''. In such cases, prepending a token solely in the reversed direction might still result in a proper reversion in semantics --- as the model could learn to associate the token with a general negation of the context (\ie ``not''). However, the same cannot be said for CIRR, where text tends to be more complicated, as shown in~\figref{fig:qualitative-0} (a-d). In such scenarios, our more balanced approach can better assist the model in identifying the directionality and associating it with said tokens.

\subsection{Qualitative Examples}

\begin{figure}[tp]
  \centering\scriptsize
  \begin{minipage}{1.00\linewidth}
    \centering
    \setlength{\tabcolsep}{0.5pt}
    \begin{tabular}{p{0.16\linewidth}ccccc}
      \textbf{(a)}& \multicolumn{5}{l}{Show three bottles of soft drink.}\\[-0.3ex]
      \textcolor{gray}{\fboxrule=1.5pt\fbox{\includegraphics[height=8.5ex]{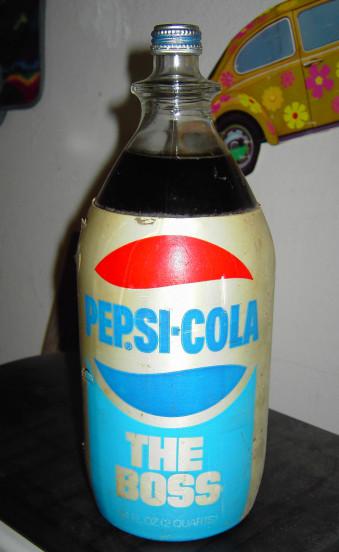}}}& 
      \frame{\includegraphics[width=11ex, height=8.5ex]{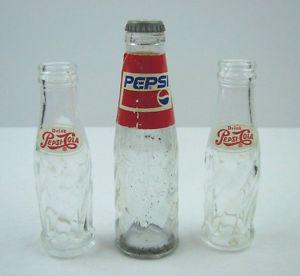}}& 
      \frame{\includegraphics[width=11ex, height=8.5ex]{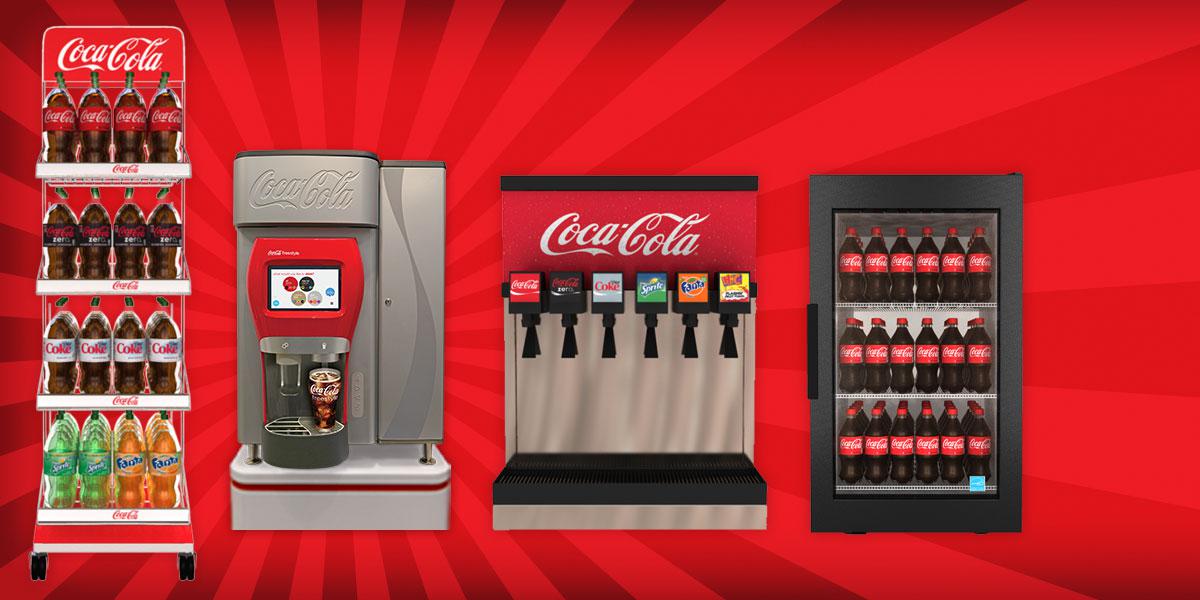}}&
      \frame{\includegraphics[width=10ex,height=8.5ex]{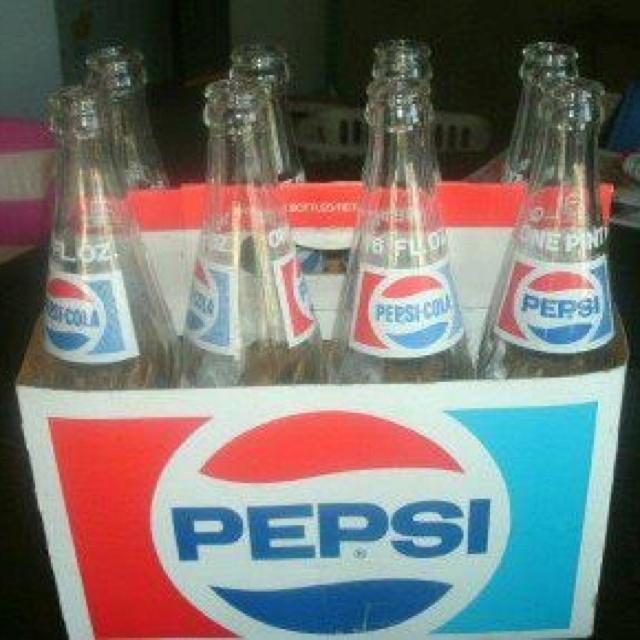}}&
      \frame{\includegraphics[height=8.5ex]{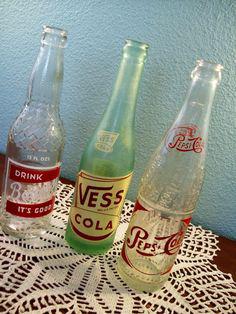}}&
      \textcolor{ForestGreen}{\fboxrule=1.5pt\fbox{\includegraphics[trim={0 0 0 0}, clip, height=8.5ex]{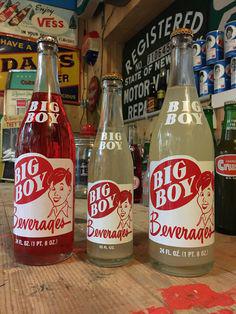}}
      }\\
      \textbf{(b)}& \multicolumn{5}{l}{Remove the humans and change the cattle to an elephant.}\\[-0.3ex]
      \textcolor{gray}{\fboxrule=1.5pt\fbox{\includegraphics[width=10.5ex, height=8.5ex]{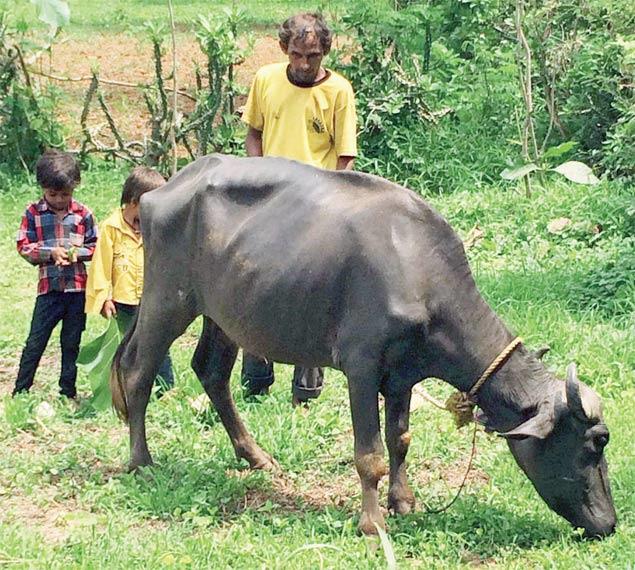}}}& 
      \frame{\includegraphics[width=10.5ex, height=8.5ex]{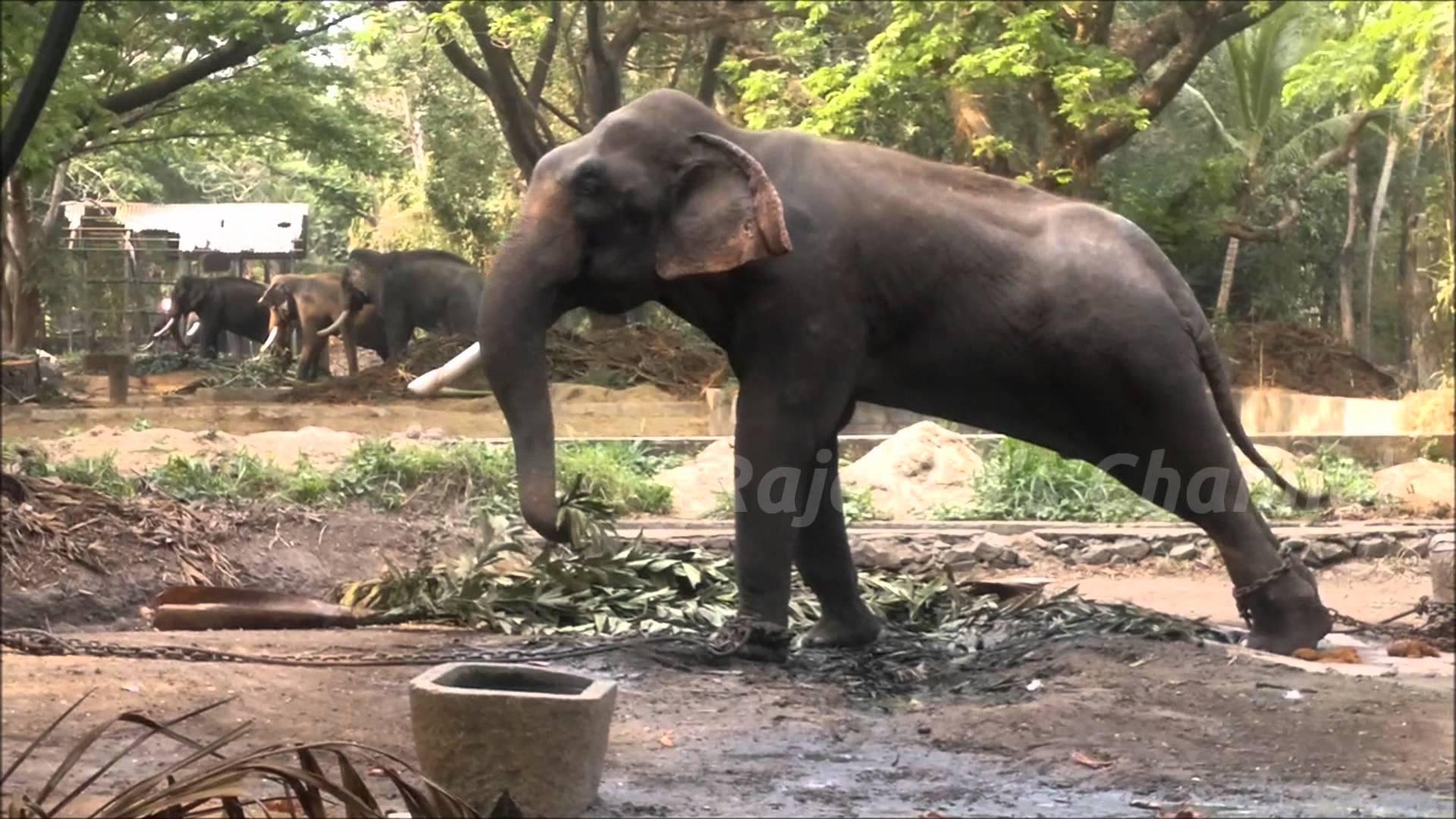}}& 
      \frame{\includegraphics[width=10.5ex, height=8.5ex]{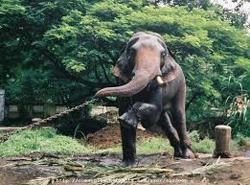}}&
      \frame{\includegraphics[width=10.5ex, height=8.5ex]{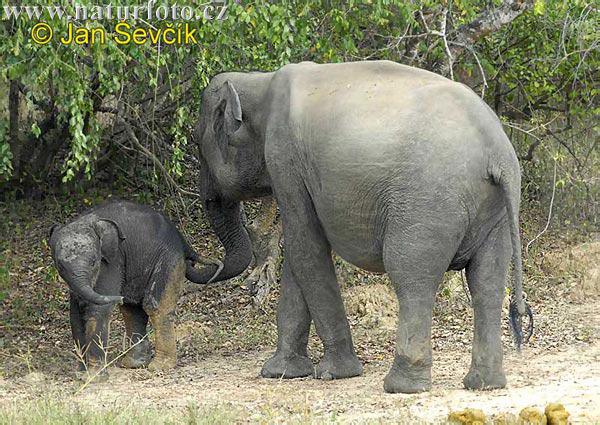}}&
      \frame{\includegraphics[width=10.5ex, height=8.5ex]{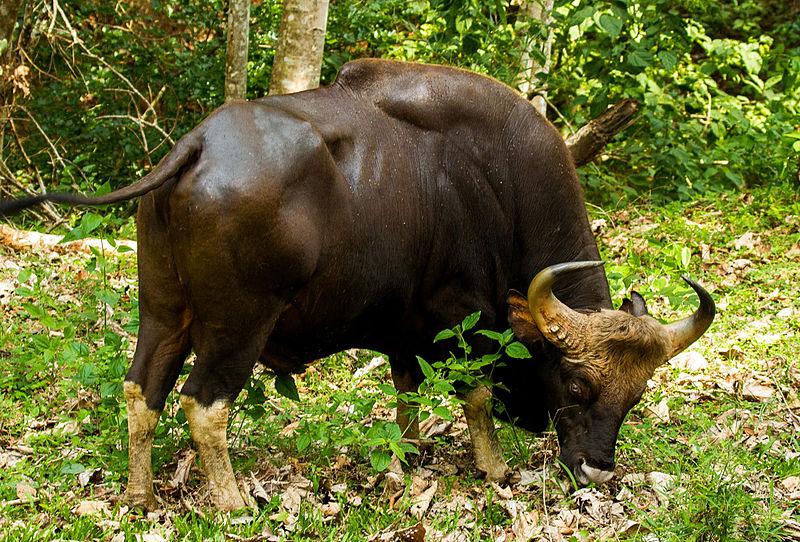}}&
      \textcolor{ForestGreen}{\fboxrule=1.5pt\fbox{\includegraphics[width=10.5ex, height=8.5ex]{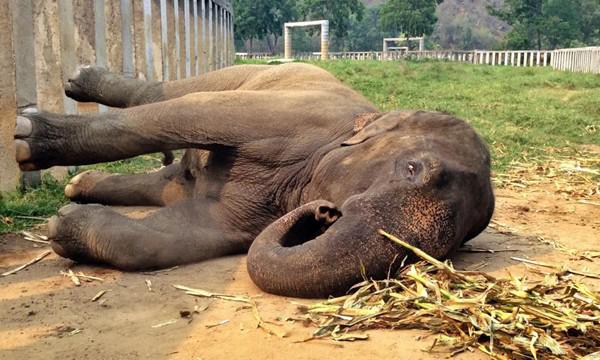}}
      }\\
      \multirow{2}{*}{\textbf{(c)}}& \multicolumn{5}{l}{Bottles must be full, no cardboard box in sight and must include}
      \\
      & \multicolumn{5}{l}{people in background.}
      \\[-0.5ex]
      \textcolor{gray}{\fboxrule=1.5pt\fbox{\includegraphics[height=8.5ex]{figs_arxiv/quali/dev-1028-3-img0}}}& 
      \frame{\includegraphics[width=10.5ex, height=8.5ex]{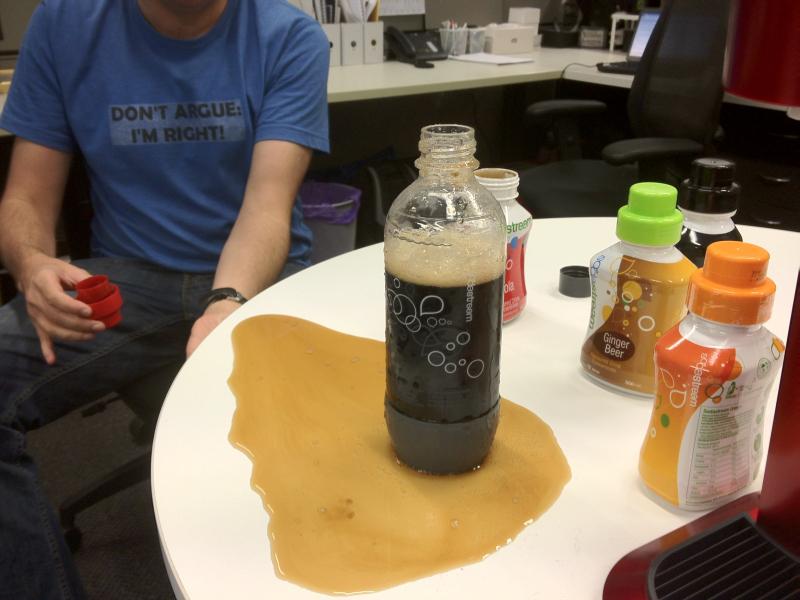}}& 
      \frame{\includegraphics[trim={0 120pt 0 100pt}, clip, height=8.5ex]{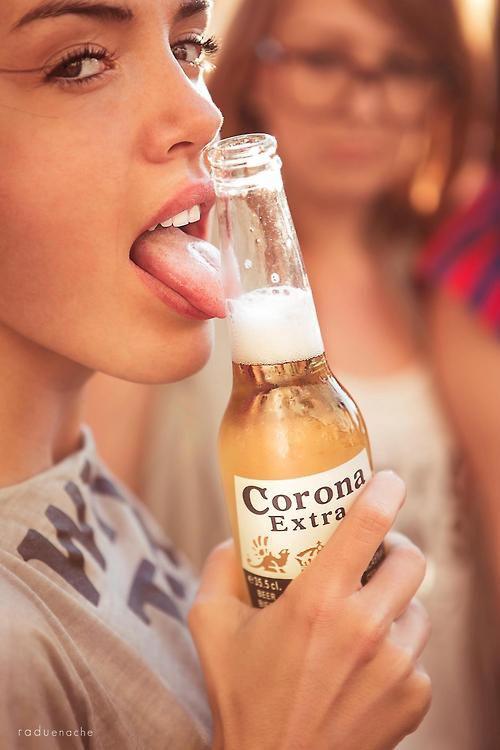}}&
      \textcolor{ForestGreen}{\fboxrule=1.5pt\fbox{\includegraphics[width=10ex, height=8.5ex]{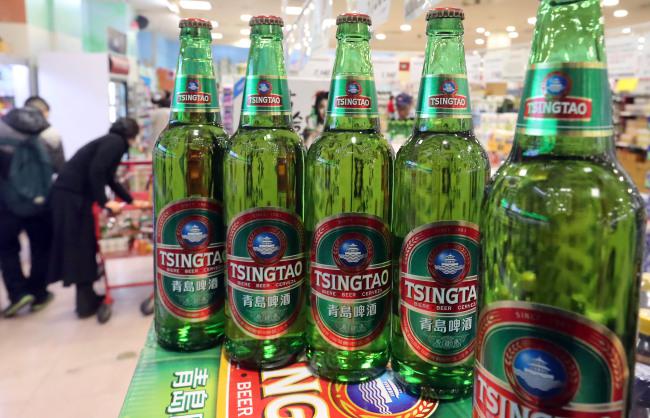}}}&
      \frame{\includegraphics[width=10.5ex, height=8.5ex]{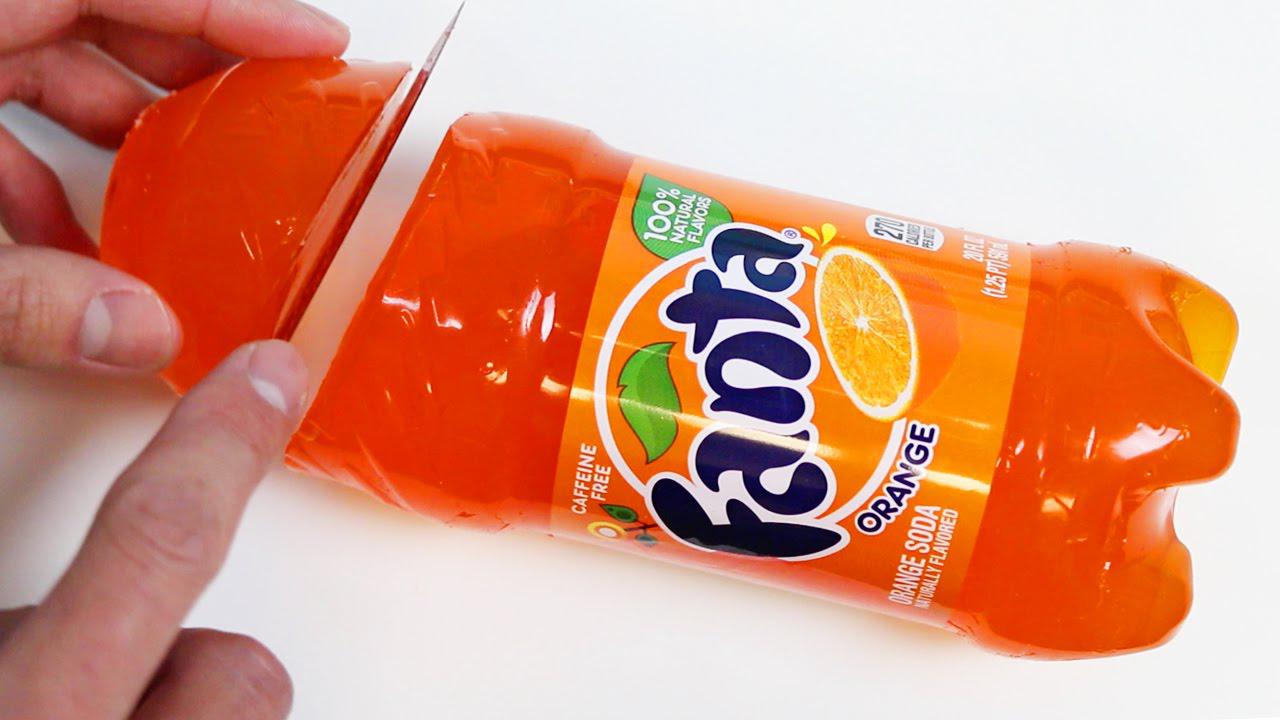}}&
      \frame{\includegraphics[width=10.5ex, height=8.5ex]{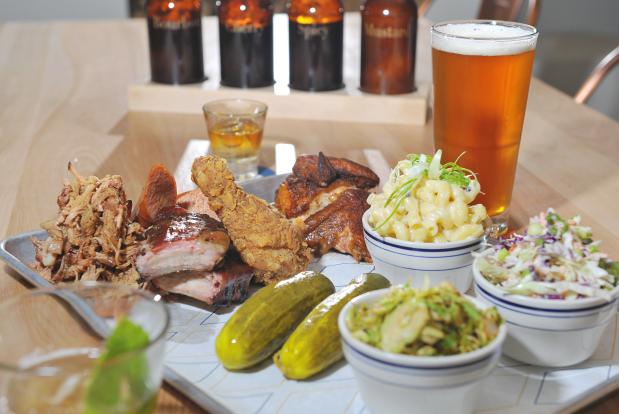}}
      \\
      
      \textbf{(d)}& \multicolumn{5}{l}{Focus on the side profile of the brown dog with blurred background.}
      \\[-0.3ex]
      \textcolor{gray}{\fboxrule=1.5pt\fbox{\includegraphics[height=8.5ex]{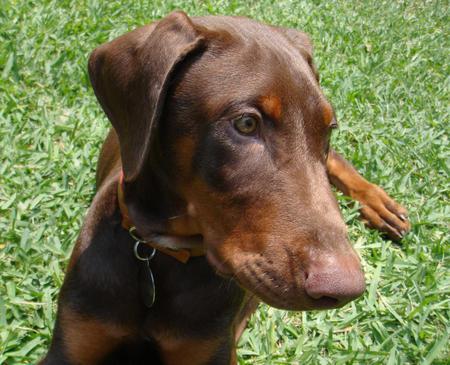}}}& 
      \textcolor{ForestGreen}{\fboxrule=1.5pt\fbox{\includegraphics[width=11ex, height=8.5ex]{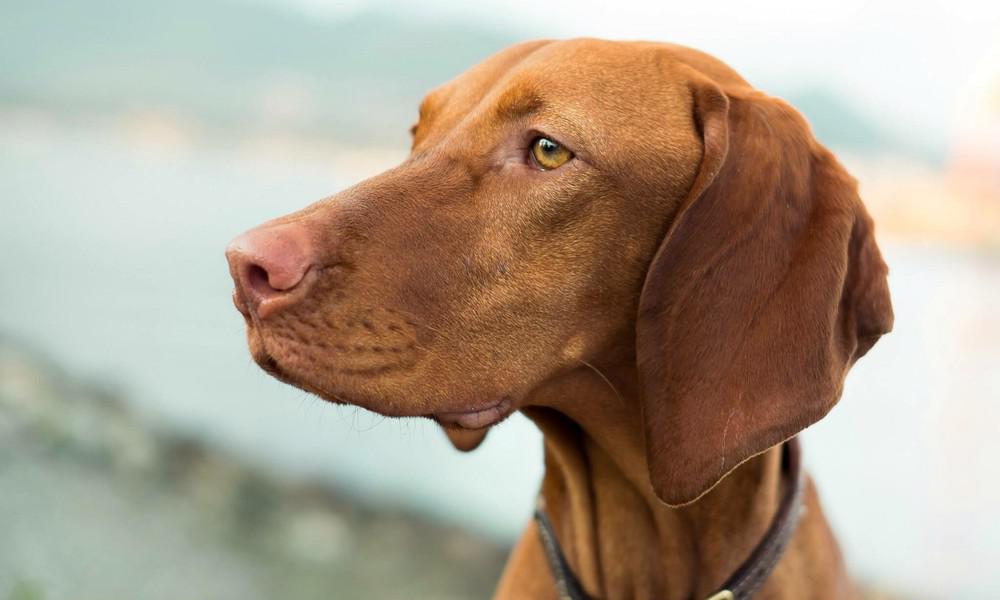}}}&
      \frame{\includegraphics[trim={0 150pt 0 150pt}, clip, height=8.5ex]{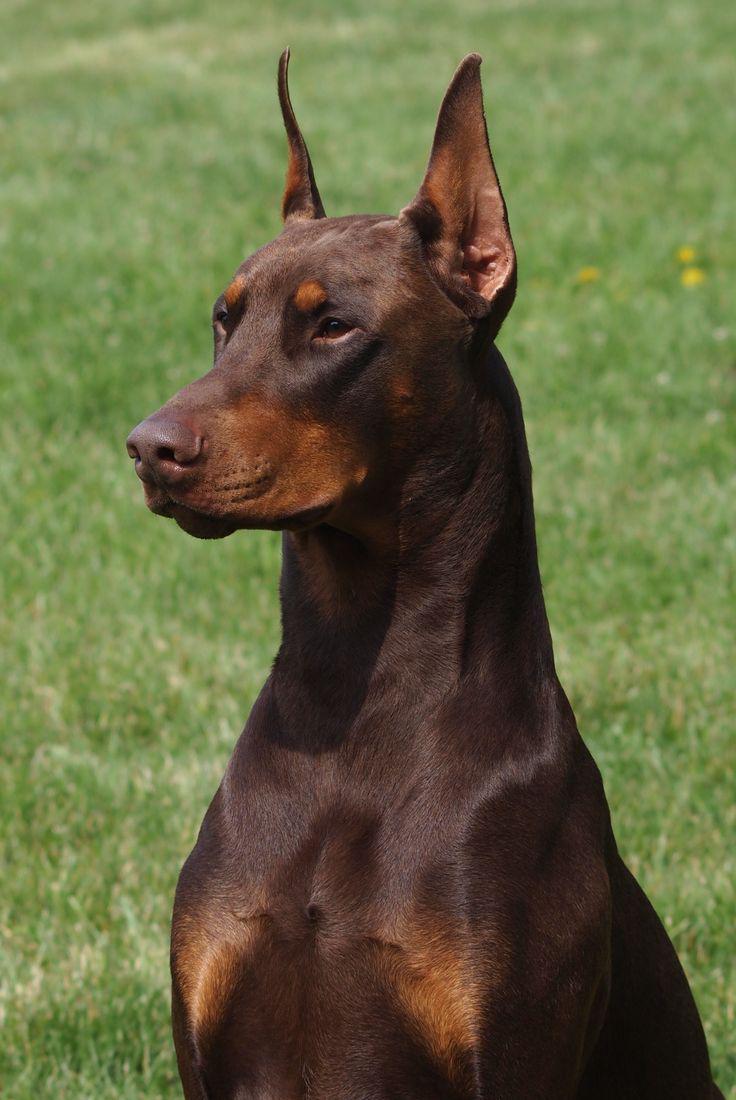}}& 
      \frame{\includegraphics[trim={0 120pt 0 30pt}, clip, height=8.5ex]{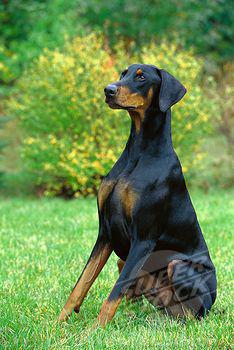}}&
      \frame{\includegraphics[width=11ex, height=8.5ex]{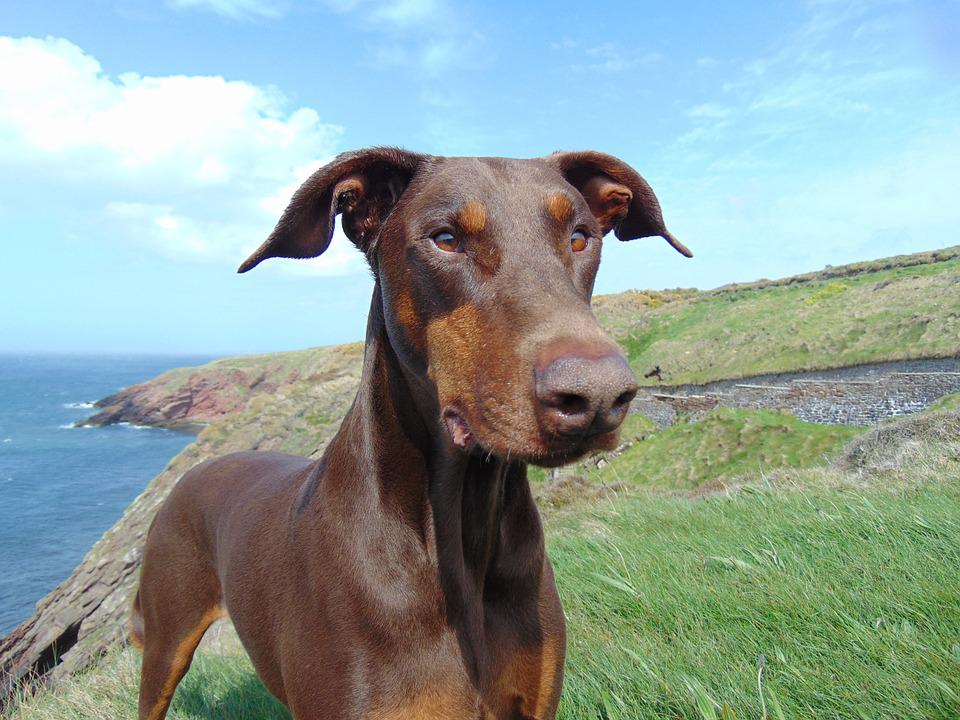}}&
      \frame{\includegraphics[trim={0 180pt 0 0}, clip, height=8.5ex]{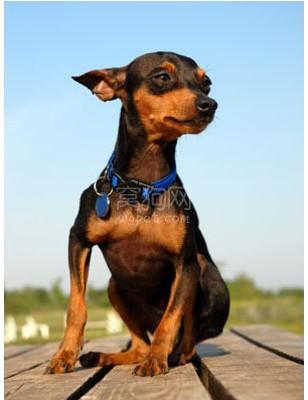}}
      \\

      \textbf{(e)}& \multicolumn{5}{l}{Is a white long sleeved shirt. Is solid white in colour.}
      \\[-0.3ex]
      \textcolor{gray}{\fboxrule=1.5pt\fbox{\includegraphics[trim={30pt 0 20pt 80pt}, clip, height=8.5ex]{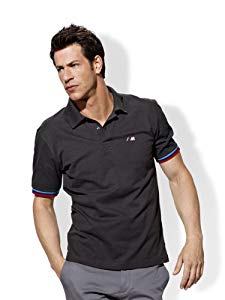}}}& 
      \frame{\includegraphics[height=8.5ex]{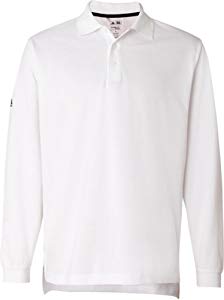}}& 
      \frame{\includegraphics[height=8.5ex]{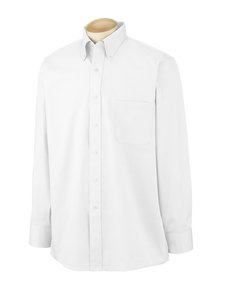}}&
      \textcolor{ForestGreen}{\fboxrule=1.5pt\fbox{\includegraphics[trim={30pt 0 20pt 80pt}, clip, height=8.5ex]{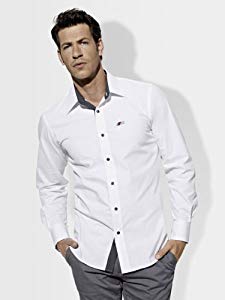}}}&
      \frame{\includegraphics[height=8.5ex]{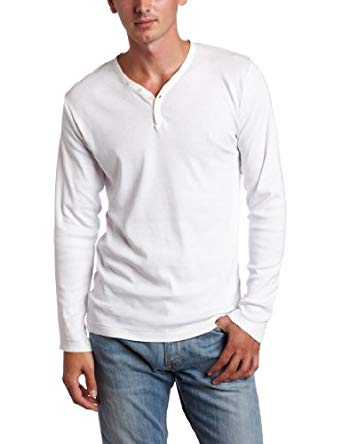}}&
      \frame{\includegraphics[height=8.5ex]{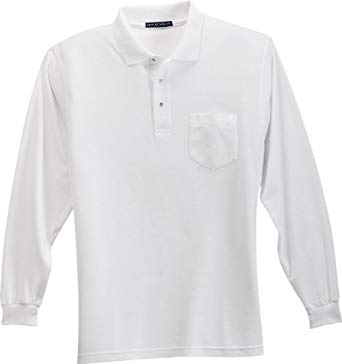}}
      \\
      \textbf{(f)}& \multicolumn{5}{l}{Is shiny and silver with shorter sleeves. Fit and flare.}
      \\[-0.3ex]
      \textcolor{gray}{\fboxrule=1.5pt\fbox{\includegraphics[height=8.5ex]{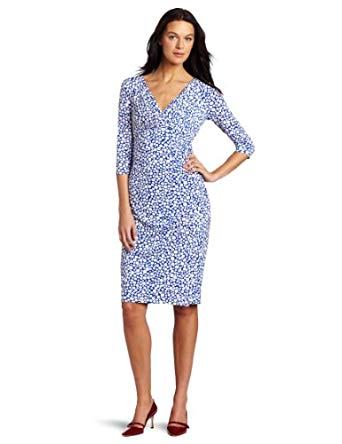}}}& 
      \frame{\includegraphics[height=8.5ex]{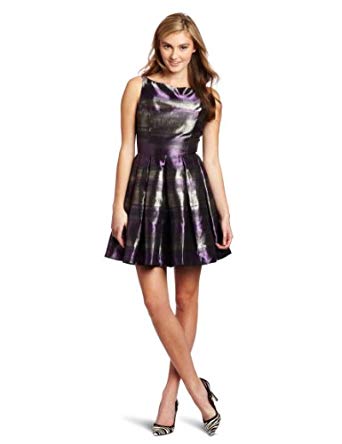}}& 
      \frame{\includegraphics[height=8.5ex]{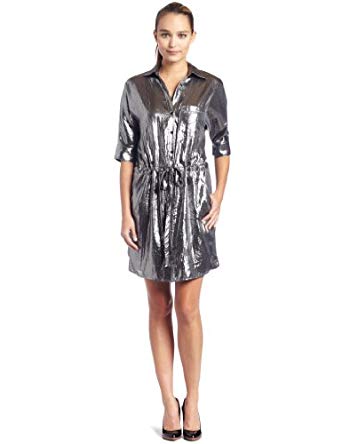}}&
      \frame{\includegraphics[height=8.5ex]{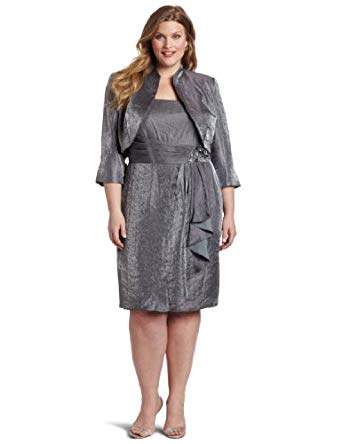}}&
      \textcolor{ForestGreen}{\fboxrule=1.5pt\fbox{\includegraphics[trim={0 0 0 0}, clip, height=8.5ex]{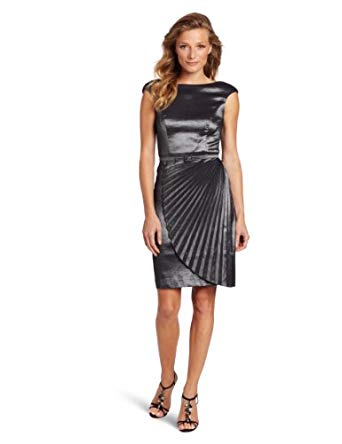}}}&
      \frame{\includegraphics[height=8.5ex]{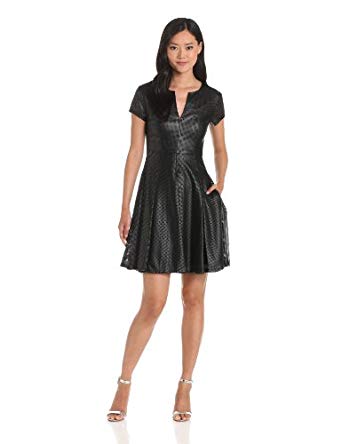}}
      \\
      \textbf{(g)}& \multicolumn{5}{l}{Has shorter sleeves. Shorter sleeved.}
      \\[-0.3ex]
      \textcolor{gray}{\fboxrule=1.5pt\fbox{\includegraphics[height=8.5ex]{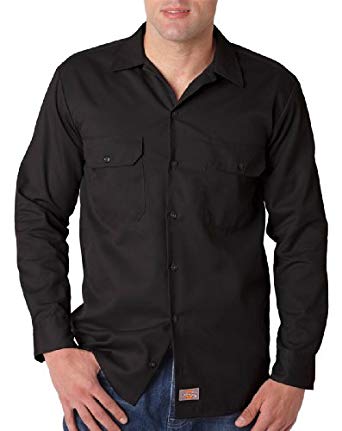}}}& 
      \frame{\includegraphics[height=8.5ex]{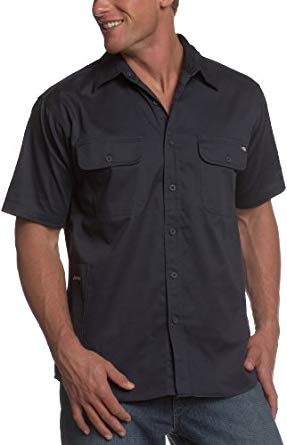}}& 
      \frame{\includegraphics[height=8.5ex]{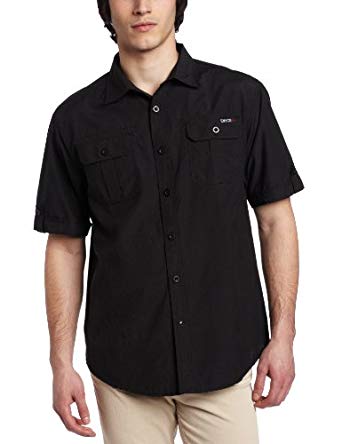}}&
      \frame{\includegraphics[height=8.5ex]{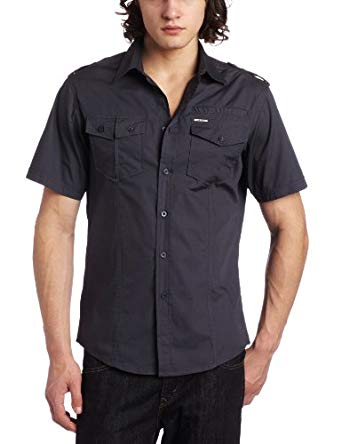}}&
      \frame{\includegraphics[height=8.5ex]{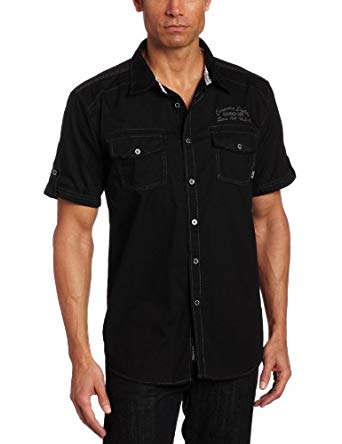}}&
      \textcolor{ForestGreen}{\fboxrule=1.5pt\fbox{\includegraphics[trim={0 0 0 0}, clip, height=8.5ex]{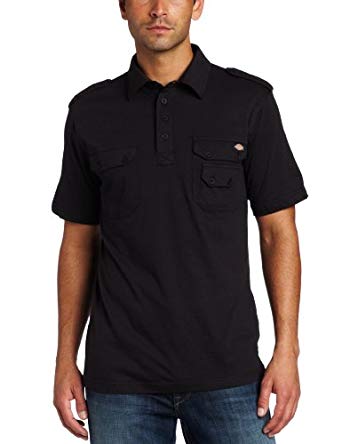}}}
      \\
      \textbf{(h)}& \multicolumn{5}{l}{Is white without collar. Is white and not coloured.}
      \\[-0.3ex]
      \textcolor{gray}{\fboxrule=1.5pt\fbox{\includegraphics[height=8.5ex]{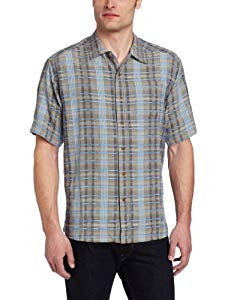}}}& 
      \frame{\includegraphics[height=8.5ex]{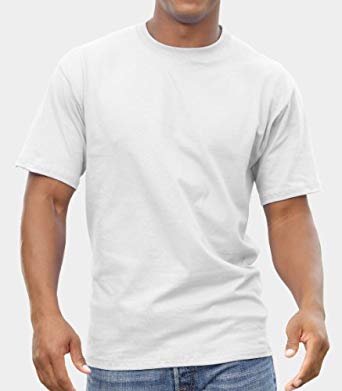}}& 
      \frame{\includegraphics[height=8.5ex]{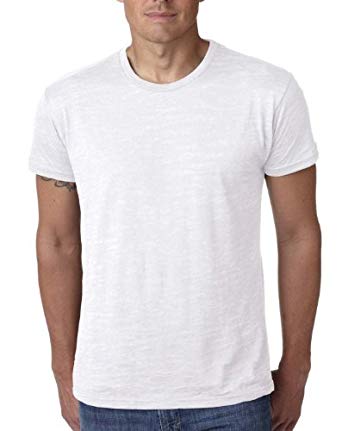}}&
      \frame{\includegraphics[height=8.5ex]{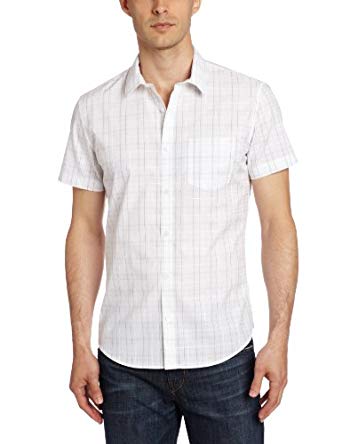}}&
      \frame{\includegraphics[height=8.5ex]{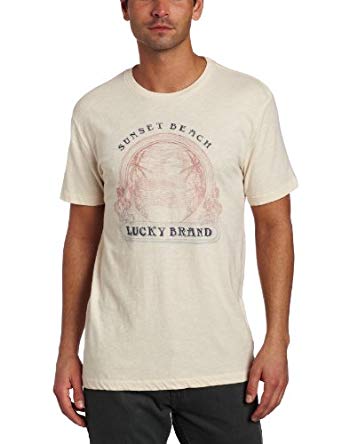}}&
      \textcolor{ForestGreen}{\fboxrule=1.5pt\fbox{\includegraphics[trim={30pt 0 30pt 0}, clip, height=8.5ex]{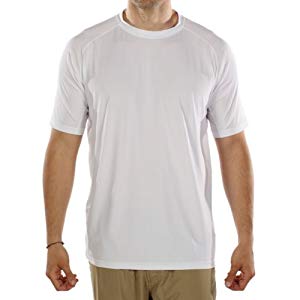}}}
    \end{tabular}
    \end{minipage}\\[-2pt]

    \caption{Qualitative examples obtained with our method.
  In each example, gray box (leftmost) denotes the reference image, green box denotes the positive target, modification text is provided above the images.
  We show the top-5 candidates in ranking, except for when the positive target is ranked beyond top-5, in which case we remove the fifth-ranked candidate and append the positive at the end. This includes examples (b), (g) and (h).
    }\label{fig:qualitative-0}
\end{figure}

\figref{fig:qualitative-0} illustrates the qualitative examples of the retrieved results on both datasets. We specifically demonstrate successful cases where positive targets are highly ranked (d, e, f), as well as failure cases (b, g, h).
We especially point to (d), where our method succeeds in reasoning over text with sophisticated intentions and retrieving the target, which demonstrates the quality of the BLIP embeddings as well as the power of our method.

As discussed in~\secref{sec:false-neg}, we expect to encounter the issue of false negatives when reversing the queries. We show that for generic, short descriptions commonly found in Fashion-IQ (e-h), the reversed text can be widely imprecise, thus, leading to a great many possible candidates. The issue is worsened as images within Fashion-IQ can be of high similarity (\eg g) --- partially due to the natural low variability of cloth images.
We show that this issue is less noticeable in CIRR, as the images are often diverse, containing multiple entities and/or rich differences in details. In addition, text in CIRR is often complex, which carries semantics that can remain specific when reversed. However, as discussed in~\secref{sec:ablation-bi-token}, the high complexity of text in CIRR also poses a challenge in the model training, as the text embeddings can be hard to semantically reverse.

\shadowoffset{2pt}
\setlength{\fboxsep}{0.2pt}

\section{Conclusion}
In this work, we propose a bi-directional training scheme for composed image retrieval that additionally exploits information from the mapping of the (target image, modification text)-pair to the reference image.
To tackle the challenge of inferring the reversed semantics of the text with the absence of additional annotations, we leverage the text encoder and prepend learnable tokens to the text inputs. Through finetuning, the text encoder binds the concept of text directionality to said tokens and can produce text embeddings for queries of either direction. We also involve a secondary contrastive loss with a modified sampling strategy for negative samples. Our approach is simple to try on any contrastive-based CIR model and can yield significant performance improvement at almost no cost. We empirically demonstrate that our bi-directional training scheme yields improved performance over a BLIP-based model that has already achieved competitive performance.

{\small
\bibliographystyle{habbrvnat}
\bibliography{egbib}
}
\clearpage
\appendix
\section*{Supplementary Material}
\section{Balancing the Forward and Reversed Loss Terms}\label{sec:balancing-losses}

We investigate the effect of varying the hyperparameter $\alpha$ in the bi-directional loss (\eqnref{eq:loss-total}).
As a general rule of thumb, we discover that large $\alpha$ values close to, or beyond 1.0 adversely harm the performance, which corroborates with our hypothesis on the effect of false negatives in the reversed direction in~\secref{sec:false-neg}. We, therefore, seek to balance the forward and reversed loss terms by reducing $\alpha$. We also note that the second-stage combiner training is more sensitive to tunings in $\alpha$ compared to the first stage. We suspect the reason to be related to the model capacity, as the first-stage finetuning is relatively light in architecture, while the second-stage combiner module is of much higher complexity (\figref{fig:model-0} right). To this end, the combiner could more easily, and quickly, overfit to the noise brought by the false negatives.

Our choices of $\alpha$ for each training stage on both datasets for results reported in Tables~\ref{tab:baseline_1} and~\ref{tab:baseline_0} are detailed as follows.
For Fashion-IQ~\cite{fashioniq}, in both stages, we discover that an $\alpha$ of around 0.5 is optimal. We note that for the first stage, further decreasing it to 0.4 yields a slightly better result. On CIRR~\cite{Liu:CIRR}, we find that the training consistently benefits from a relatively small $\alpha$, we set it to 0.1 in both stages.

\begin{figure}[th]
  \begin{center}
    \includegraphics[trim={0pt 0pt 0pt 0pt},clip, width=0.49\linewidth]{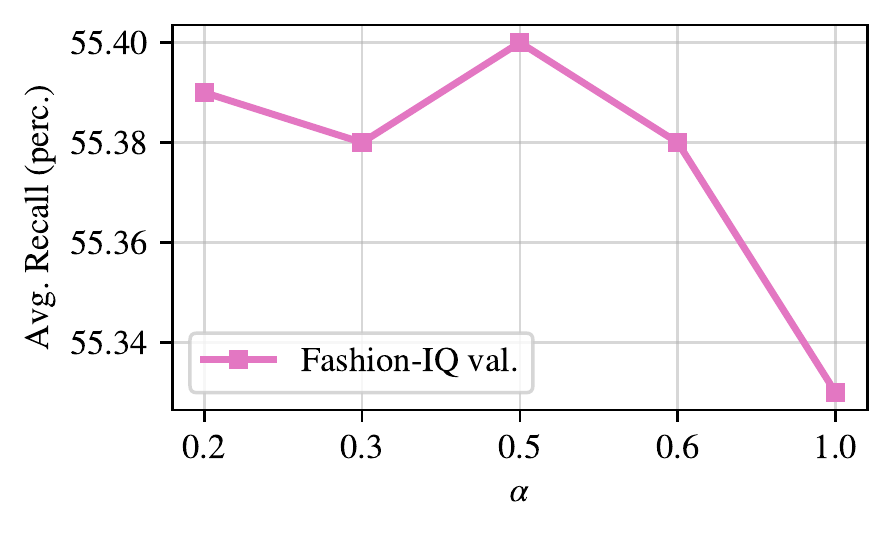}
    \includegraphics[trim={0pt 0pt 0pt 0pt},clip, width=0.48\linewidth]{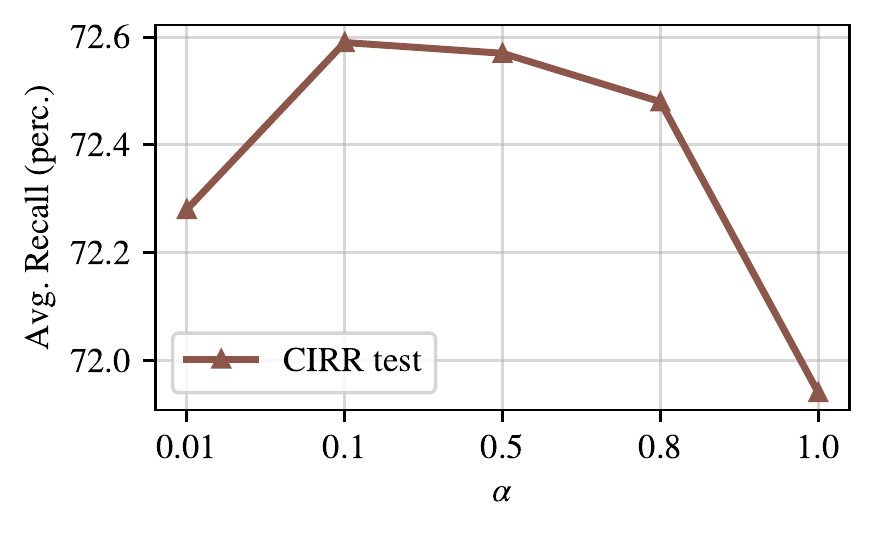}
  \end{center}
  \caption[Performance \textit{vs.} $\alpha$ in the second stage bi-directional training]{Performance \textit{vs.} $\alpha$ in the second stage bi-directional training. \textbf{(Left)} Fashion-IQ validation set. \textbf{(Right)} CIRR test set. We select a few $\alpha$ values to examine the trend surrounding optimality. Note the relatively small scale in performance (y-axis), suggesting the performance is fairly robust against varying $\alpha$ within a certain range. Compare the results with Tables~\ref{tab:baseline_1} and~\ref{tab:baseline_0}.
  }\label{fig:alpha_ablation}
  \vspace{-7pt}
\end{figure}

In \figref{fig:alpha_ablation} we illustrate the effect of varying $\alpha$ on performance in the second-stage combiner training. We notice that as long as $\alpha$ sits within a certain range that is smaller than 1.0, the results are fairly robust. 

\section{Inference on Reversed Queries}\label{sec:inference-on-reversed-queries}

\secref{sec:false-neg} details the impact of false-negatives. In~\tabref{tab:val-reversed} we demonstrate that validating on the reversed queries yields subpar results, which collaborates with our observation of a higher loss in the reversed path. This leads to our inference strategy that only takes into account the forward queries.

\begin{table*}[tp]
  \centering \scalebox{0.7}{
  \begin{tabular}{p{0.03\linewidth}p{0.2\linewidth}rrrrrrr} 
  \toprule
   &    & \multicolumn{3}{c}{\textbf{Fashion-IQ}}               & \multicolumn{4}{c}{\textbf{CIRR}} \\
  \cmidrule(lr){3-5}
  \cmidrule(lr){6-9}
  \multicolumn{1}{l}{} & \textbf{BLIP4CIR+Bi} & R@10  & R@50  & Average & R@1 &  R@5   & R$_\text{Subset}$@1  & Average  \\ 
  \midrule
  \textbf{1}  &  on forward queries  & 43.49 & 67.31 & 55.40 & 42.36 & 75.46  & 72.90 & 74.18 \\ 
  \textbf{2}  &  on reversed queries  & 23.08 & 45.05 & 34.07 & 18.08 & 49.25  & 44.51 & 46.88 \\ 
  \bottomrule
  \end{tabular}}
  \caption{Comparison of performance when validating on the forward and reversed queries. Results obtained on validation sets after the second-stage combiner training, directly comparable to results in~\tabref{tab:ablate_0}.
  }\label{tab:val-reversed}
\end{table*}

\begin{table*}[tp]
  \centering
  \scalebox{0.7}{
  \begin{tabular}{p{0.03\linewidth}lrrrrrrrrr} 
  \toprule
  \multicolumn{1}{c}{} & \multicolumn{1}{c}{} & \multicolumn{2}{c}{\textbf{Dress}} & \multicolumn{2}{c}{\textbf{Shirt}} &\multicolumn{2}{c}{\textbf{Toptee}} &\multicolumn{2}{c}{\textbf{Average}} & \textbf{Avg.} \\
  \cmidrule(lr){3-4}
  \cmidrule(lr){5-6}
  \cmidrule(lr){7-8}
  \cmidrule(lr){9-10}
  \multicolumn{1}{l}{} & \multicolumn{1}{l}{\textbf{Methods}} & R@10 & R@50 & R@10 & R@50 & R@10 & R@50 & R@10 & R@50 & \textbf{Metric}  \\ 
  \midrule
  \textbf{1} & BLIP4CIR~(first-stage) & 4.81 & 15.42 & 8.10 & 16.63  & 7.75 & 17.64  & 6.89 & 16.57 & 11.72   \\ 
  \textbf{2} & BLIP4CIR+Bi~(first-stage) & 22.91 & 45.96 & 23.80 & 41.22  & 27.03 & 45.44  & 24.58 & 44.20 & 34.39  \\ 
  \bottomrule
  \end{tabular}}
  \caption[Performance comparison on the reversed query retrieval with or without bi-directional training on Fashion-IQ validation set]{Performance comparison on the \textit{reversed query} retrieval with or without bi-directional training, Fashion-IQ validation set. We report the average Recall@10 and 50 of all three categories. Note that the comparison is on the first-stage text encoder finetuning (\figref{fig:model-0} left).}
  \label{tab:main1-fiq_reversed_val}
\end{table*}

\setlength{\fboxsep}{0.75pt}
\begin{figure*}[tp]
  \centering\scriptsize
  \begin{minipage}{0.99\linewidth}
    \centering
    \setlength{\tabcolsep}{1.5pt}
    \begin{tabular}{p{0.17\linewidth}ccccc}
      \textbf{(a)}& \multicolumn{5}{l}{\texttt{[BACKWARD]} Add one more deer and add some sunlight.}
      \\[.8ex]
      \frame{\includegraphics[height=12.5ex]{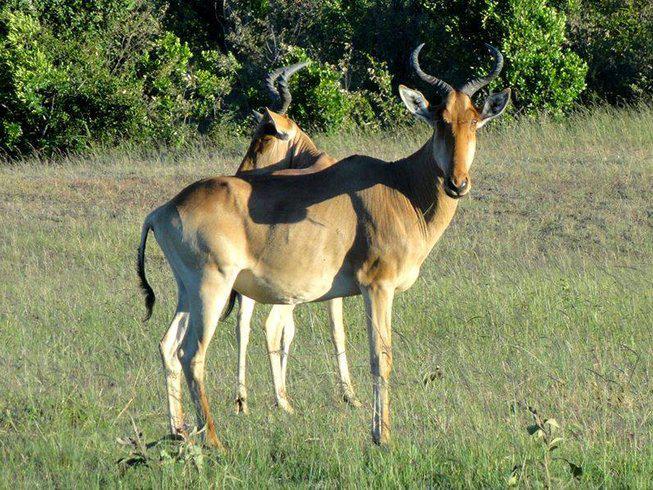}}& 
      \textcolor{ForestGreen}{\fboxrule=2pt\fbox{\includegraphics[trim={0 0 0 0}, clip, width=14ex, height=12.5ex]{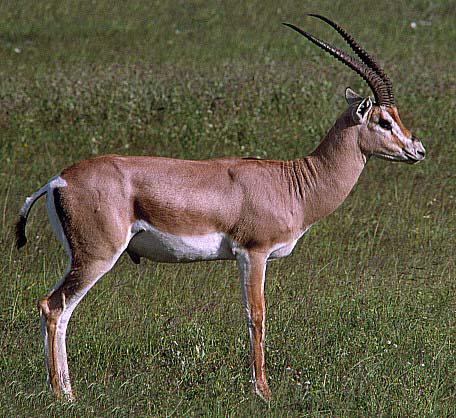}}}&
      \frame{\includegraphics[width=14ex, height=12.5ex]{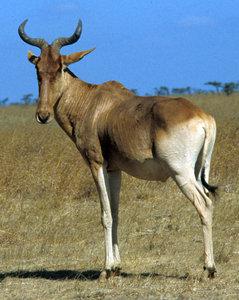}}&
      \frame{\includegraphics[width=14ex, height=12.5ex]{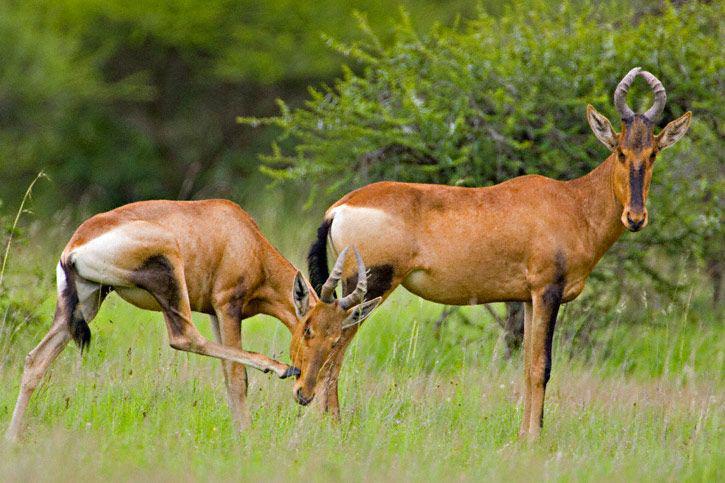}}& 
      \frame{\includegraphics[width=8ex, height=12.5ex]{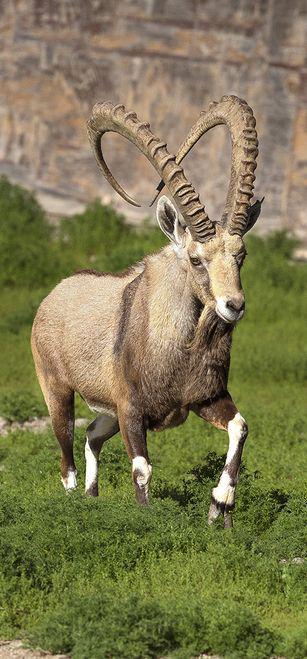}}&
      \frame{\includegraphics[width=14ex, height=12.5ex]{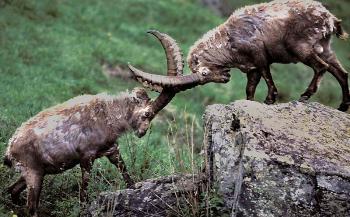}}
      \\[2ex]
      
      \textbf{(b)}& \multicolumn{5}{l}{\texttt{[BACKWARD]} Put the fries in a white plate with white background, clean.}
      \\[.8ex]
      \frame{\includegraphics[height=12.5ex]{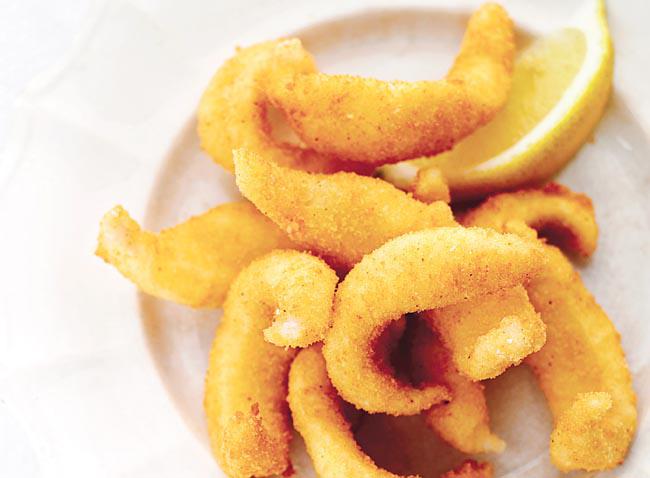}}& 
      \textcolor{ForestGreen}{\fboxrule=2pt\fbox{\includegraphics[width=15ex, height=12.5ex]{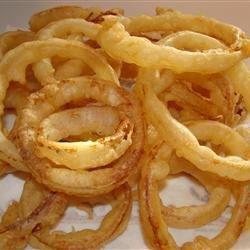}}}&
      \frame{\includegraphics[width=16ex, height=12.5ex]{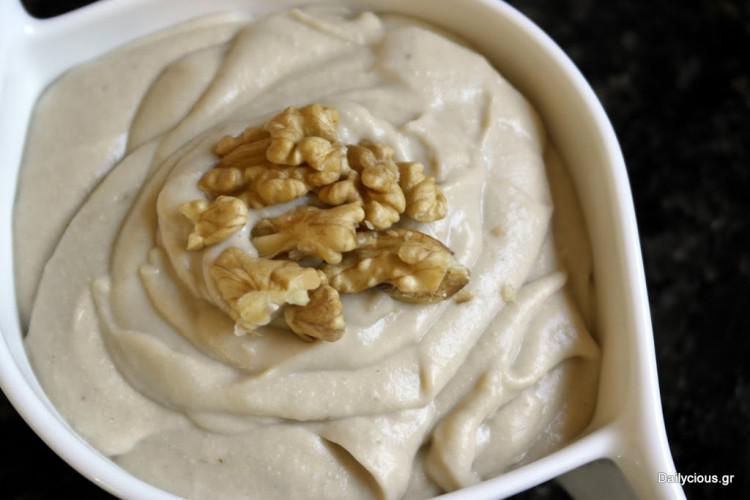}}& 
      \frame{\includegraphics[height=12.5ex]{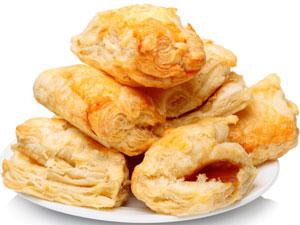}}&
      \frame{\includegraphics[width=15ex, height=12.5ex]{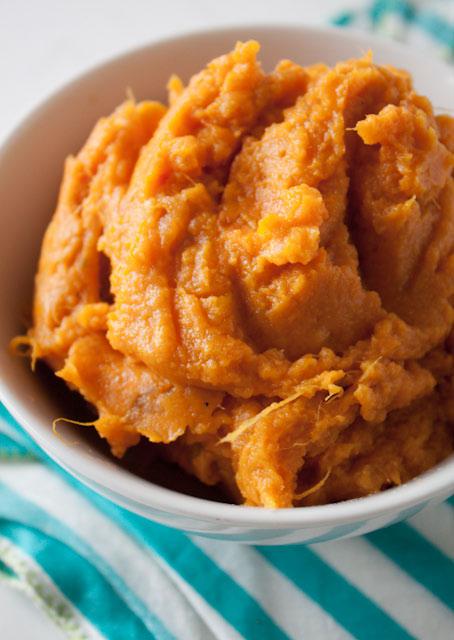}}&
      \frame{\includegraphics[width=16ex, height=12.5ex]{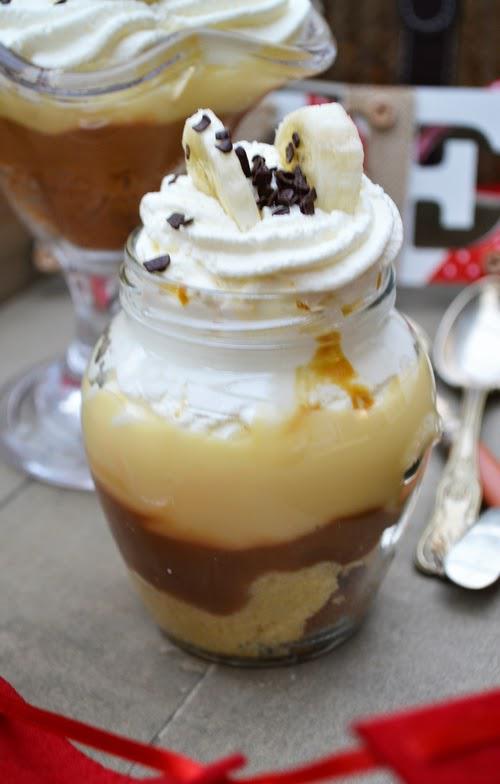}}
      \\[2ex]
      
      \textbf{(c)}& \multicolumn{5}{l}{\texttt{[BACKWARD]} Change the plate to rectangular.}
      \\[.8ex]
      \frame{\includegraphics[trim={0 0 0 0}, clip, height=12.5ex]{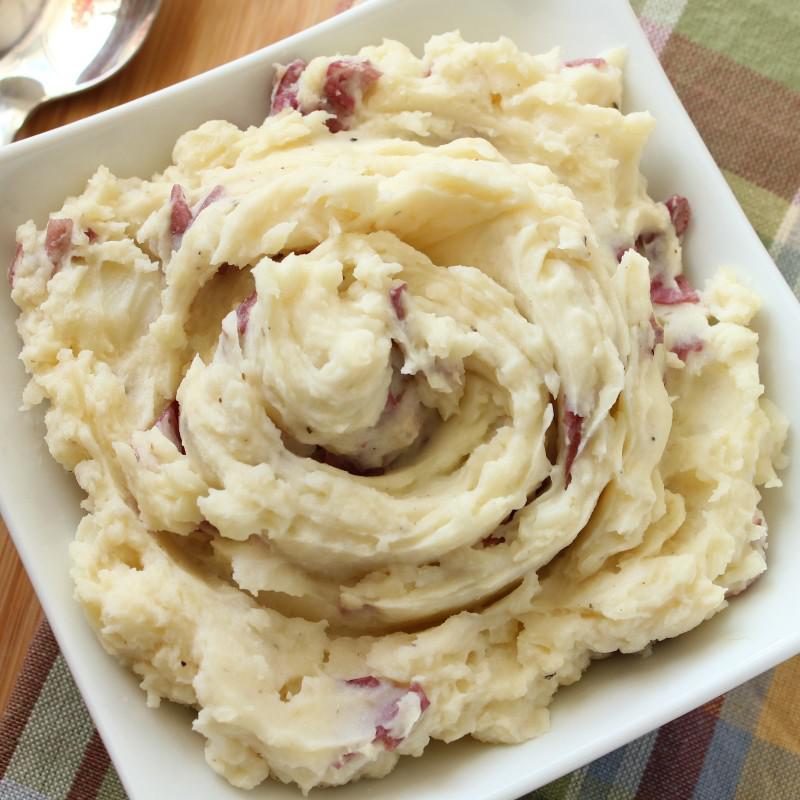}}& 
      \frame{\includegraphics[height=12.5ex]{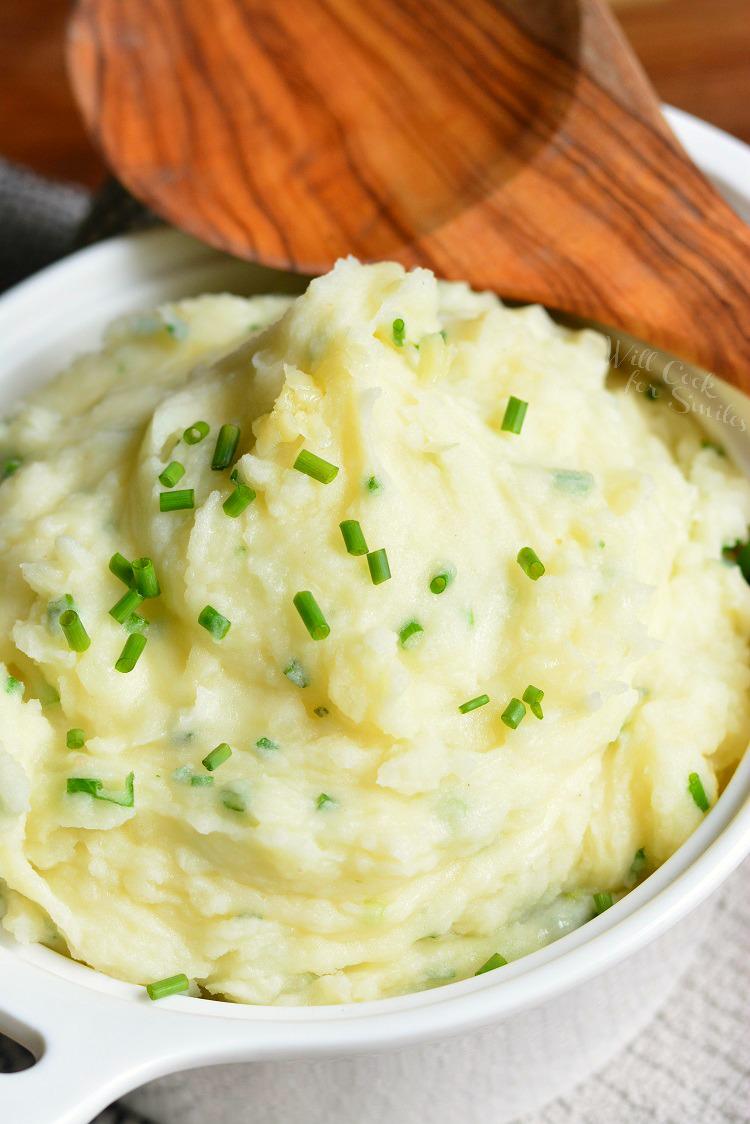}}& 
      \textcolor{ForestGreen}{\fboxrule=2pt\fbox{\includegraphics[width=16ex, height=12.5ex]{figs_arxiv/quali/dev-187-2-img0}}}&
      \frame{\includegraphics[trim={0 0 0 0}, clip, height=12.5ex]{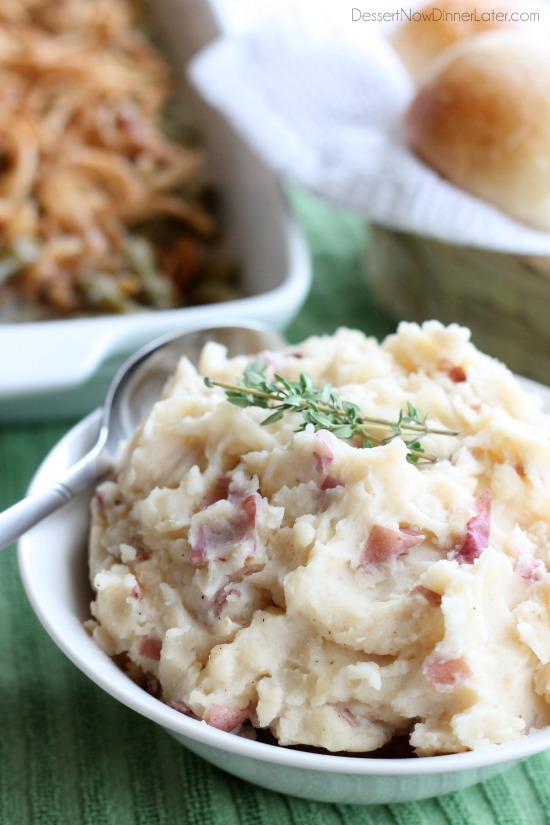}}&
      \frame{\includegraphics[trim={0 0 0 0}, clip, width=16ex,  height=12.5ex]{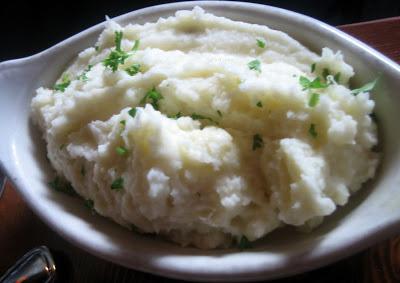}}&
      \frame{\includegraphics[width=16ex, height=12.5ex]{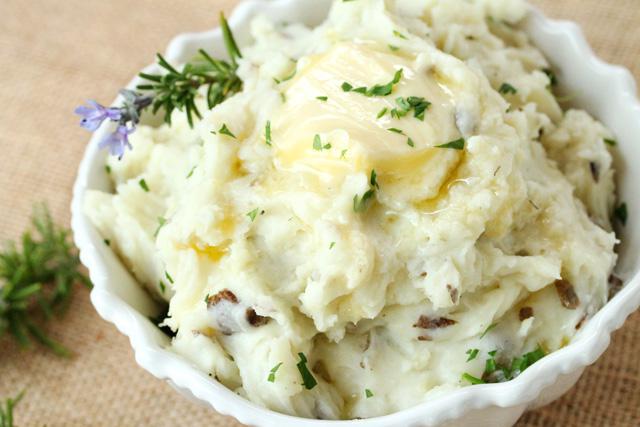}}
      \\[2ex]
      
      \textbf{(d)}& \multicolumn{5}{l}{\texttt{[BACKWARD]} Fewer paper towels per pack.}
      \\[.8ex]
      \frame{\includegraphics[width=12ex, height=12.5ex]{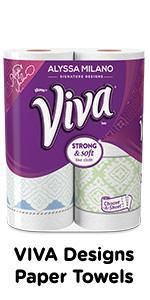}}& 
      \textcolor{ForestGreen}{\fboxrule=2pt\fbox{\includegraphics[width=14ex, height=12.5ex]{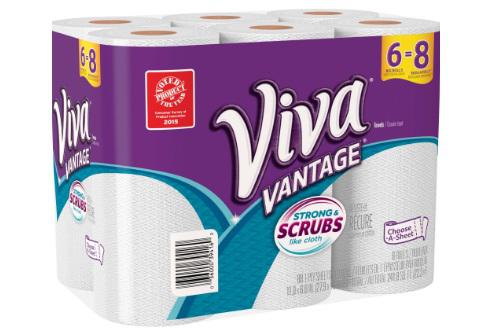}}}&
      \frame{\includegraphics[width=16ex, height=12.5ex]{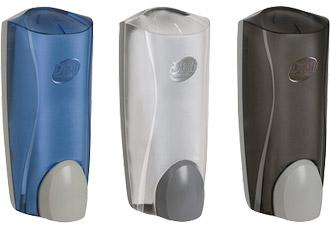}}&
      \frame{\includegraphics[width=9ex, height=12.5ex]{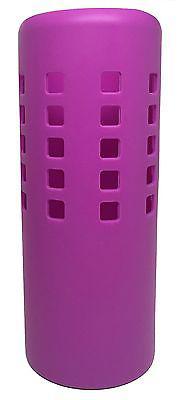}}&
      \frame{\includegraphics[width=16ex, height=12.5ex]{figs_arxiv/quali/dev-1028-1-img1}}&
      \frame{\includegraphics[width=16ex, height=12.5ex]{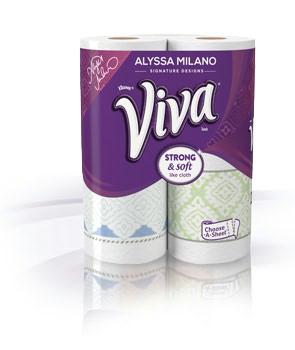}}
      \\[2ex]
      
    \end{tabular}
    \end{minipage}\\[5pt]

    \caption[Qualitative examples of reversed query retrieval on CIRR]{Qualitative examples of \textit{reversed query} retrieval on the first-stage text encoder finetuning (CIRR). 
    In each example, leftmost is target image, green box denotes the ground truth (reference image), the reversed modification text is provided above the images. We show the top-5 candidates in ranking.
    Note that the reference image and target image exchange roles here and that the modification text shall be interpreted in its reversed semantic --- for this, we specifically show the prepended text token.
  }\label{fig:main1-qualitative-reversed-cirr}
\end{figure*}

\section{Analysis on the Learned Reversed Semantics}\label{sec:analysis-on-not-tokens}
We perform both quantitative and qualitative analyses to examine if our bi-directional training is encouraging the learning of the reversed semantics.
Specifically, \tabref{tab:main1-fiq_reversed_val} (rows 2 \textit{vs.} 1) compares the retrieval performance on the Fashion-IQ reversed queries with or without bi-directional training.
We examine the model after the first-stage text encoder finetuning, as in~\figref{fig:model-0} (left).
The result suggests that a model specifically trained with bi-directional queries is better equipped at reasoning over reversed semantics, which substantiates our claim.
However, note that the performance on said queries is generally much lower than on the (standard) forward ones due to the larger number of potential false negatives, which has been discussed in~\secref{sec:false-neg}.

We additionally present four qualitative examples of CIRR retrieved on the reversed queries.
In~\figref{fig:main1-qualitative-reversed-cirr} (a) and (d) where the reversed text is unambiguous (\ie ``add'' is negated to ``remove'', ``fewer'' is negated to ``more''), we show the model is capable of reasoning over such reversed semantics. 
We demonstrate a more complicated case in (b), where one might not definitively predict the ground truth content by examining the query. Still, among the top-5 ranked candidates, we argue that the model produces a plausible result, with the ground truth ranked the highest.

We especially illustrate the existence of false negatives among candidates in~\figref{fig:main1-qualitative-reversed-cirr} (c) --- though the issue is present in multiple examples. 
Here, in particular, ``change to rectangular'' shall be reversed to ``change \textit{from} rectangular'', which points to a range of possible shapes. Indeed, the top-5 ranked candidates all contain non-rectangular plates --- though only one of them is labelled positive. 
Here, we note that not all such reversed examples with false negatives can be successfully retrieved.  Evidence can be seen when comparing the performance on the reversed queries (\tabref{tab:main1-fiq_reversed_val} row 2) to the performance on the forward ones (\tabref{tab:baseline_1} row 19), where the former is much lower than the latter.
This further validates our decisions to not perform inference on the reversed queries (\secref{sec:false-neg}) and to downscale the reversed contrastive loss (\secref{sec:balancing-losses}).

\end{document}